\documentclass[twoside,11pt]{article}
\usepackage{jair, theapa, rawfonts}

\usepackage{url}
\usepackage{color}
\usepackage{comment}
\usepackage{graphicx}
\usepackage{booktabs}

\ShortHeadings{A Survey on Vector Representations of Meaning}
{Camacho-Collados \& Pilehvar}
\firstpageno{1}

\begin{document}

\title{From Word to Sense Embeddings: \\ A Survey on Vector Representations of Meaning}


\author{\name Jose Camacho-Collados \email camachocolladosj@cardiff.ac.uk \\
       \addr School of Computer Science and Informatics \\
       Cardiff University \\ United Kingdom \\
       \AND
       \name Mohammad Taher Pilehvar \email pilehvar@iust.ac.ir \\
       \addr School of Computer Engineering \\Iran University of Science and Technology \\ Tehran, Iran
       }


\maketitle

\begin{abstract}

Over the past years, distributed semantic representations have proved to be effective and flexible keepers of prior knowledge to be integrated into downstream applications. 
This survey focuses on the representation of meaning.
We start from the theoretical background behind word vector space models and highlight one of their major limitations: the meaning conflation deficiency, which arises from representing a word with all its possible meanings as a single vector. 
Then, we explain how this deficiency can be addressed through a transition from the word level to the more fine-grained level of word senses (in its broader acceptation) as a method for modelling unambiguous lexical meaning. 
We present a comprehensive overview of the wide range of techniques in the two main branches of sense representation, i.e., unsupervised and knowledge-based. 
Finally, this survey covers the main evaluation procedures and applications for this type of representation, and provides an analysis of four of its important aspects: interpretability, sense granularity, adaptability to different domains and compositionality. 

\end{abstract}

\section{Introduction}
\label{Introduction}

Recently, neural network based approaches which process massive amounts of textual data to embed words' semantics into low-dimensional vectors, the so-called word embeddings, have garnered a lot of attention \cite{Mikolovetal:2013,pennington2014glove}. 
Word embeddings have demonstrated their effectiveness in storing valuable syntactic and semantic information \cite{mikolov2013linguistic}.
In fact, they have been shown to be beneficial to many Natural Language Processing (NLP) tasks, mainly due to their generalization power \cite{goldberg2016primer}. 
A wide range of applications have reported improvements upon integrating word embeddings, including machine translation \cite{zou2013bilingual}, syntactic parsing \cite{weiss2015structured}, text classification \cite{kim2014convolutional} and question answering \cite{bordes2014question}, to name a few. 

However, despite their flexibility and success in capturing semantic properties of words, the effectiveness of word embeddings is generally hampered by an important limitation which we will refer to as \textit{meaning conflation deficiency}: the inability to discriminate among different meanings of a word.
A word can have one meaning (monosemous) or multiple meanings (ambiguous).  For instance, the noun \textit{nail} can refer to two different meanings depending on the context: a part of the finger or a metallic object. Hence, the noun \textit{nail} is said to be ambiguous\footnote{\textit{Nail} can also refer to a unit of cloth measurement (generally a sixteenth of a yard) or even be used as a verb.}.
Each individual meaning of an ambiguous word is called a word sense and a lexical resource that lists different meanings (senses) of words is usually referred to as a sense inventory.\footnote{In order to obtain the list of possible word senses of a target word, lexicographers tend to first collect occurrences of the words from corpora and then manually cluster them semantically and based on their contexts, i.e., \textit{concordance} \cite{Kilgarriff:97b}. Given this procedure, \citeauthor{Kilgarriff:97b} \citeyear{Kilgarriff:97b} suggested that word senses, as defined by sense inventories in NLP, should not be construed as objects but rather as abstractions over clusters of word usages.} While most words in general sense inventories (e.g. WordNet) are often monosemous\footnote{For instance, around 83\% of the 155K words in WordNet 3.0 are listed as monosemous (see Section \ref{inventories} for more information on lexical resources).}, frequent words tend to have more senses, according to the Principle of Economical Versatility of Words \cite{Zipf:49}. Therefore, accurately capturing the semantics of ambiguous words plays a crucial role in the language understanding of NLP systems.

In order to deal with the meaning conflation deficiency, a number of approaches have attempted to model individual word senses. 
In this survey we have tried to synthesize the most relevant works on sense representation learning. 
The main distinction of these approaches is in how they model meaning and where they obtain it from. 
\textit{Unsupervised} models directly learn word senses from text corpora, while \textit{knowledge-based} techniques exploit the sense inventories of lexical resources as their main source for representing meanings. 
In this survey we cover these two classes of techniques for learning distributed semantic representations of meaning, including evaluation procedures and an analysis of their main properties. 
While the survey is intended to be as extensive as possible, given the breadth of the topics reviewed, some areas may not have received a sufficient coverage to be totally self-contained. However, for these cases we provide relevant pointers for readers interested in learning more on the topic. Given the wide audience that this survey is intended to reach, we have tried to make it as understandable as possible. Therefore, technical details might not have been necessarily provided in full detail, but rather the intuition behind them.

The remainder of this survey is structured as follows. First, in Section \ref{theoretical} we provide a theoretical background for word senses, what they are, why modeling them may be useful and its main paradigms. Then, in Section \ref{unsupervised} we describe unsupervised sense vector modeling techniques which learn directly from text corpora, while in Section \ref{knowledgebased} the representations linked to lexical resources are explained. Common evaluation procedures and benchmarks are presented in Section \ref{sec:evaluation} and the applications in downstream tasks in Section \ref{applications}. Finally, we present an analysis and comparison between unsupervised and knowledge-based representations in Section \ref{analysis} and the main conclusions and future challenges in Section \ref{conclusion}.




\section{Background}
\label{theoretical}

This section provides theoretical foundations which support the move from word level to the more fine-grained level of word senses and concepts. First, we provide the background to vector space models, particularly for word representation learning (Section \ref{wordrepresentationlearning}). Then, we explain some of the main deficiencies of word representations 
which led to the development of sense modeling techniques (Section \ref{conflation}) 
and describe the main paradigms for representing senses (Section \ref{paradigms}). In Section \ref{wsd} we present a brief historical background of the related task of word sense disambiguation. Finally, we explain the notation followed throughout the survey (Section \ref{notation}).


\subsection{Word Representation Learning}
\label{wordrepresentationlearning}

Word representation learning has been one of the main research areas in Semantics since the beginnings of NLP. We first introduce the main theories behind word representation learning based on vector space models (Section \ref{vsm}) and then move to the emerging theories for learning word embeddings (Section \ref{wordembeddings}). 

\subsubsection{Vector Space Models}
\label{vsm}

One of the most prominent methodologies for word representation learning is based on Vector Space Models (VSM), which is supported by research in human cognition \cite{Landauer:1997,gardenfors2004conceptual}. The earliest VSM applied in NLP considered a document as a vector whose dimensions were the whole vocabulary \cite{Saltonetal:1975}. Weights of individual dimensions were initially computed based on word frequencies within the document. Different weight computation metrics have been explored, but mainly based on frequencies or normalized frequencies \cite{SaltonMcGill:83}. This methodology has been successfully refined and applied to various NLP applications such as information retrieval \cite{lee1997document}, text classification \cite{soucy2005beyond}, or sentiment analysis \cite{turney2002thumbs}, to name a few. \citeauthor{TurneyPantel:10} \citeyear{TurneyPantel:10} provide a comprehensive overview of VSM and their applications.

\begin{figure}[t!]
\centering
    \includegraphics[trim = 0mm 0mm 0mm 0mm,,scale=0.22
    ]{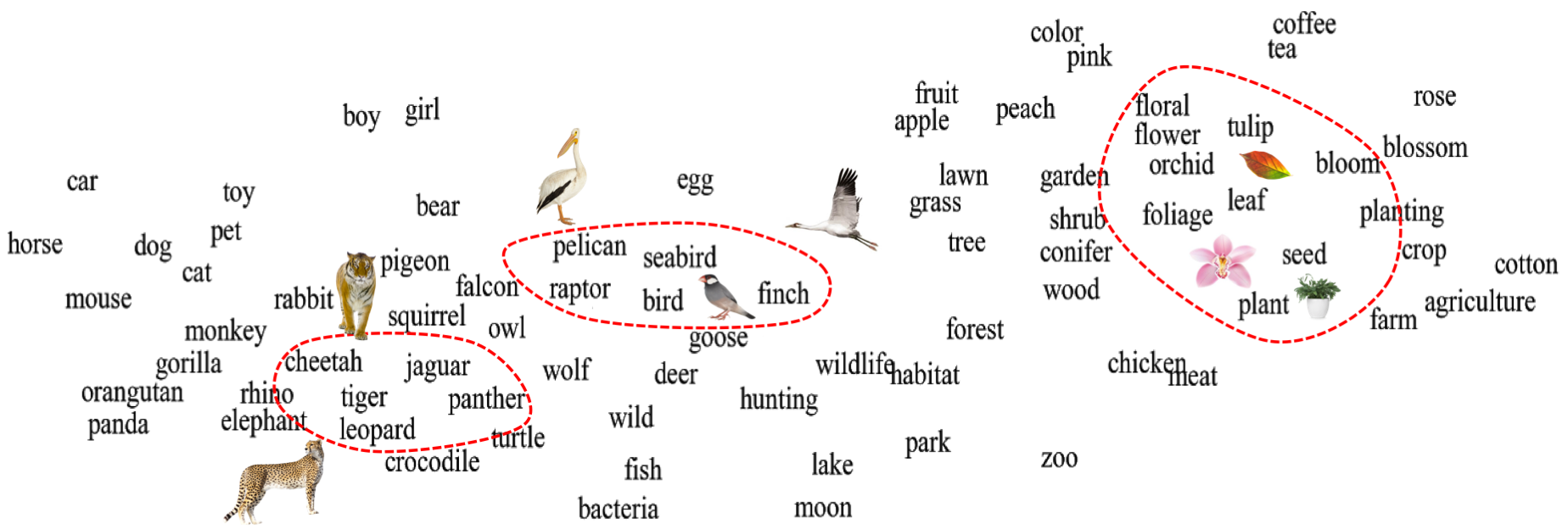}
    \caption{Subset of a sample word vector space reduced to two dimensions using t-SNE
 \cite{maaten2008visualizing}. In a semantic space, words with similar meanings tend to appear in the proximity of each other, as highlighted by these word clusters (delimited by the red dashed lines) associated with \textit{big cats}, \textit{birds} and \textit{plants}.}
    \label{fig:tsneBigCats}
\end{figure}

The document-based VSM has been also extended to other lexical items like words. 
In this case a word is generally represented as a point in a vector space. A word-based vector has been traditionally constructed based on the normalized frequencies of the co-occurring words in a corpus \cite{lund1996producing}, by following the initial theories of \citeauthor{Harris:54} \citeyear{Harris:54}. The main idea behind word VSM is that words that share similar context should be close in the vector space (therefore, have similar semantics). 
Figure \ref{fig:tsneBigCats} shows an example of a word VSM where this underlying proximity axiom is clearly highlighted.

Vector-based representations have established their effectiveness in NLP tasks such as information extraction \cite{Laenderetal:2002}, semantic role labeling \cite{Erk:07}, word similarity \cite{Radinskyetal:2011}, word sense disambiguation \cite{navigli:09} or spelling correction \cite{JonesMartin:1997}, \textit{inter alia}.
One of the main drawbacks of the conventional VSM approaches is the high dimensionality of the produced vectors. Since the dimensions correspond to words in the vocabulary, this number could easily reach hundreds of thousands or even millions, depending on the underlying corpus. 
A common approach for dimensionality reduction makes use of the Singular Value Decomposition (SVD), also known as Latent Semantic Analysis \cite[LSA]{hofmann2001unsupervised,LandauerDooley:2002}. In addition, recent models also leverage neural networks to directly learn low-dimensional word representations. These models are introduced in the following section.  


\subsubsection{Word Embeddings}
\label{wordembeddings}

Learning low-dimensional vectors from text corpora can alternatively be achieved by exploiting neural networks. These models are commonly known as \textit{word embeddings} and have been shown to provide valuable prior knowledge thanks to their generalization power \cite{goldberg2016primer}. This property has proved to be decisive for achieving state-of-the-art performance in many NLP tasks when integrated into a neural network architecture \cite{zou2013bilingual,kim2014convolutional,bordes2014question,weiss2015structured}.

\begin{figure}[t!]
\centering
    \includegraphics[trim = 0mm 0mm 0mm 0mm,,scale=0.17
    ]{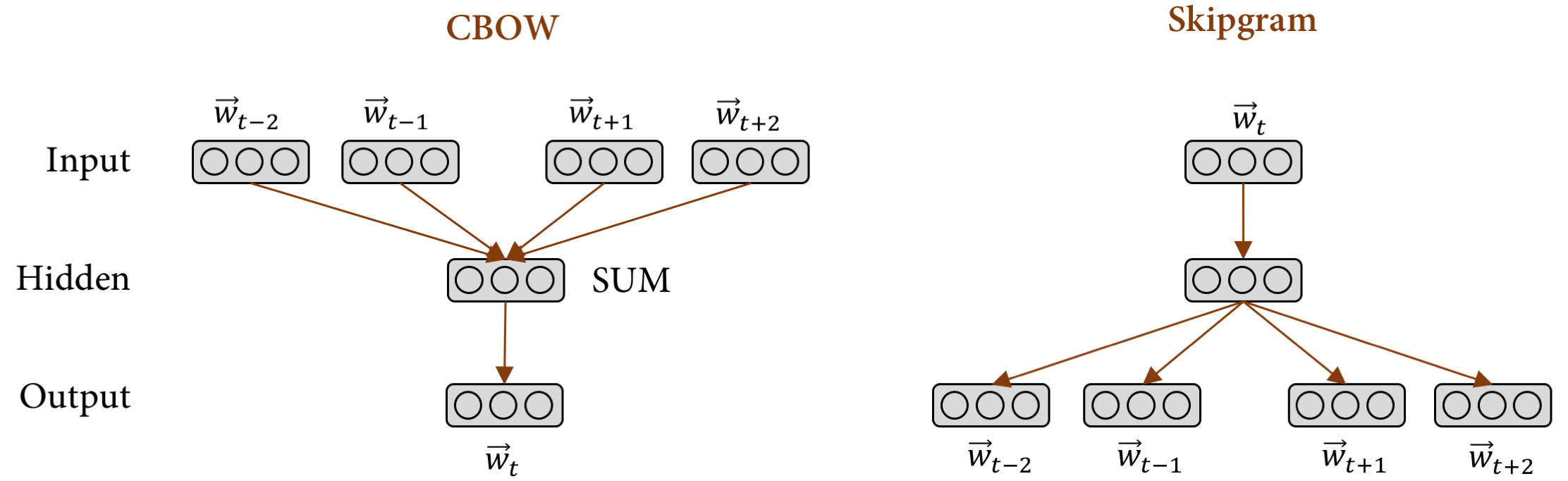}
    \caption{Learning architecture of the CBOW and Skipgram models of Word2vec \cite{Mikolovetal:2013}.}
    \label{fig:Word2vec}
\end{figure}

This newer predictive branch, whose architecture is based on optimizing a certain objective \cite{bengio2003neural,Collobert:2008,Turianetal:2010,Collobert:2011}, was popularized through Word2vec \cite{Mikolovetal:2013}. Word2vec is based on a simple but efficient architecture which provides interesting semantic properties \cite{mikolov2013linguistic}. Two different but related Word2vec models were proposed: Continuous Bag-Of-Words (CBOW) and Sikp-gram. The CBOW architecture is based on a feedforward neural network language model \cite{bengio2003neural} and aims at predicting the current word using its surrounding context, minimizing the following loss function:

\begin{equation}
E=-\log(p(\vec{w_{t}} | \vec{W_t}))
\end{equation}


\noindent where $w_{t}$ is the target word and $W_t=w_{t-n},...,w_t,...,w_{t+n}$ represents the sequence of words in context. Figure \ref{fig:Word2vec} shows a simplification of the general architecture of the CBOW and Skip-gram models of Word2vec. The architecture consists of input, hidden and output layers. The input layer has the size of the word vocabulary and encodes the context as a combination of one-hot vector representations of surrounding words of a given target word. The output layer has the same size as the input layer and contains a one-hot vector of the target word during the training phase. The Skip-gram model is similar to the CBOW model but in this case the goal is to predict the words in the surrounding context given the target word, rather than predicting the target word itself. Interestingly, \citeauthor{levy2014neural} \citeyear{levy2014neural} proved that Skip-gram can be in fact viewed as an implicit factorization of a Point-Mutual Information (PMI) co-occurrence matrix. 

Another prominent word embedding architecture is GloVe \cite{pennington2014glove}, which combines global matrix factorization and local context window methods through a bilinear regression model. In recent years more complex approaches that attempt to improve the quality of word embeddings 
have been proposed, including models exploiting dependency parse-trees \cite{levy2014dependency} or symmetric patterns \cite{schwartz2015symmetric}, leveraging subword units \cite{wieting2016charagram,bojanowski2017enriching}, representing words as probability distributions \cite{vilnisword2015,athiwaratkun2017multimodal,athiwaratkun2018multisense}, learning word embeddings in multilingual vector spaces \cite{conneau2018word,artetxeacl2018unsupervised}, or exploiting knowledge resources (more details about this type in Section \ref{wordknowledgerep}).\footnote{For a more comprehensive overview on word embeddings and their current challenges, please refer to the work of \citeauthor{ruder2017word} \citeyear{ruder2017word}.} 


\subsection{Meaning Conflation Deficiency}
\label{conflation}

\begin{figure}[t!]
\centering
    \includegraphics[trim={0 4cm 0 0},width=0.6\textwidth]{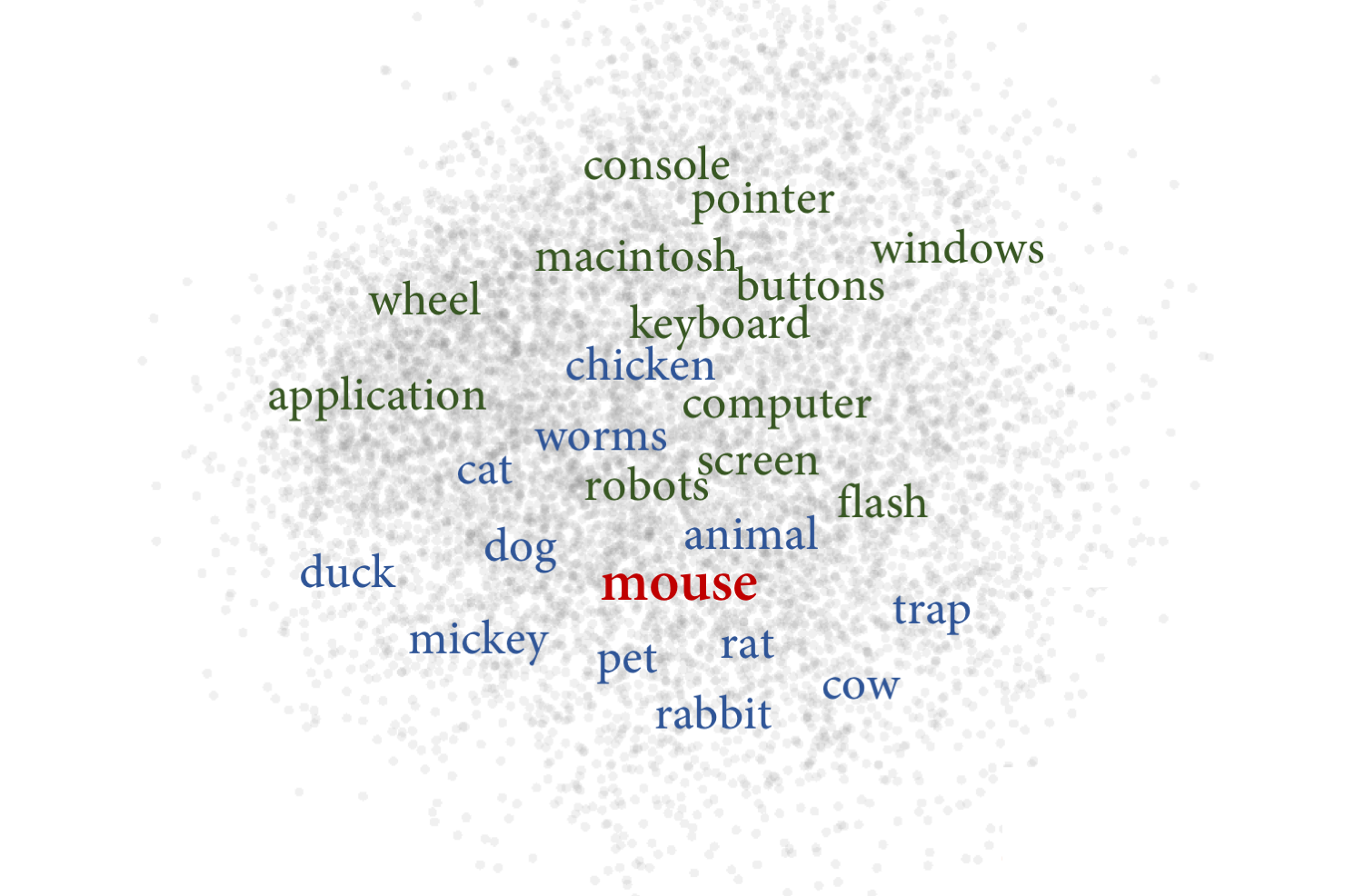}
    \caption{An illustration of the meaning conflation deficiency in a 2D semantic space around the ambiguous word \textit{mouse}.
    Having the word, with its different meanings, represented as a single point (vector) results in pulling together of semantically unrelated words, such as \textit{computer} and \textit{rabbit}.}
    \label{fig:conflation}
\end{figure}

The prevailing objective of representing each word type as a single point in the semantic space has a major limitation: it ignores the fact that words can have multiple meanings and conflates all these meanings into a single representation. The work of \citeauthor{schutze1998automatic} \citeyear{schutze1998automatic} is one of the earliest to identify the meaning conflation deficiency of word vectors. Having different (possibly unrelated) meanings conflated into a single representation can hamper the semantic understanding of an NLP system that uses these at its core. In fact, word embeddings have been shown to be unable in effectively capturing different meanings of a word, even when these meanings occur in the underlying training corpus \cite{Yaghoobzadeh:acl2016}. The meaning conflation can have additional negative impacts on accurate semantic modeling, e.g., semantically unrelated words that are similar to different senses of a word are pulled towards each other in the semantic space \cite{Neelakantanetal:2014,PilehvarCollier:2016emnlp}. 
For example, the two semantically-unrelated words \textit{rat} and \textit{screen} are pulled towards each other in the semantic space for their similarities to two different senses of \textit{mouse}, i.e., rodent and computer input device.
See Figure \ref{fig:conflation} for an illustration.\footnote{Dimensionality was reduced using PCA; visualized with http://projector.tensorflow.org/.}
Moreover, the conflation deficiency violates the triangle inequality of euclidean spaces, which can reduce the effectiveness of word space models \cite{TverskyGate:1982}. 
In order to alleviate this deficiency, a new direction of research has emerged over the past years, which tries to directly model individual meanings of words.
In this survey we focus on this new branch of research, which has some similarities and peculiarities with respect to word representation learning.

\subsection{Sense Representation}
\label{paradigms}

A solution to addressing the meaning conflation deficiency of word embeddings is to represent individual meanings of words, i.e., word senses, as independent representations.
Such representations are generally referred to as sense representations.
Sense representation techniques can be broadly classified depending on the way sense distinctions are made.
The optimal way of partitioning the meanings of words into multiple senses has long been the point of argument \cite{erk2009investigations,erk2012vector,mccarthy2016word}.
Traditionally, as in word sense disambiguation (see Section \ref{wsd}), computational techniques have relied on fixed sense inventories produced by humans, such as WordNet \cite{miller1995wordnet}.
A sense inventory\footnote{In Section \ref{inventories} we provide an overview of few of the most popular sense inventories.} is a lexical resource, such as a dictionary or thesaurus, that lists for each word the possible meanings it can take.
Sense distinctions can also be defined through word sense induction, i.e., automatic identification of a word's senses by analyzing the contexts in which it appears.

Generally, sense representations can be divided into two main paradigms depending on how the sense distinctions are defined:

\begin{itemize}
\item {\bf Unsupervised.} In these representation models the sense distinctions are induced by analyzing text corpora. This paradigm is very related to word sense induction.

\item {\bf Knowledge-based.} These techniques represent word senses as defined by an external sense inventory, e.g., WordNet.\footnote{In this survey we also cover representations directly linked to knowledge resources even if senses are not explicitly listed (e.g., concepts and entities in Wikipedia), including knowledge base embeddings.} In this case, the most directly associated task is word sense disambiguation. 

\end{itemize}

In Sections \ref{unsupervised} and \ref{knowledgebased}, we will provide details of each paradigm and their variants.








\subsection{Word Sense Disambiguation}
\label{wsd}

Word Sense Disambiguation (WSD) is a task which is closely related to the meaning conflation deficiency. WSD has been a long-standing task in NLP and AI \cite{navigli:09}, dating back to the first half of the 20th century where it was viewed as a key intermediate task for machine translation \cite{weaver1955translation}. Given a word in context, the task of WSD consists of associating the word with its most appropriate meaning as defined by a sense inventory. For example, in the sentence ``My \textit{mouse} was broken, so I bought a new one yesterday.", \textit{mouse} would be associated with its \textit{computer device} meaning, assuming an existing entry for such sense in the pre-defined sense inventory.

WSD has been catalogued as an AI-complete problem \cite{Mallery:88,navigli:09} and its challenges (still present nowadays) are manifold: sense granularity, corpus domain or the representation of word senses (topic addressed in this survey), to name a few. In addition, the fact that WSD relies on knowledge resources poses additional challenges such as the creation of such resources and the construction of sense-annotated corpora. All of these represent a very expensive and time-consuming effort, which needs to be re-done for different resources and languages, and updated over time. This causes the so-called knowledge-acquisition bottleneck \cite{Galeetal:92b}.

The knowledge resources and sense inventories traditionally used in WSD have been associated with entries on a standard dictionary, with WordNet \cite{Milleretal:93} being the de-facto sense inventory for WSD. Nevertheless, other machine-readable structures can be (and are) considered in practice. For example, Wikipedia, which is constantly being updated, can be viewed as a sense inventory where each entry corresponds to a different concept or entity \cite{mihalcea:07b}. Senses can even be induced automatically from a corpus using unsupervised methods, a task known as word sense induction or discrimination.

Methods to perform WSD can be roughly divided into two classes: supervised \cite{ZhongNg:2010,iacobacci-pilehvar-navigli:2016:P16-1,yuan2016word,raganato2017neural,Luo2018glosses} and knowledge-based \cite{Lesk:86,BanerjeePedersen:2002,agirre2014random,Moroetal:14tacl,tripodi2017game,chaplot2018knowledge}. While supervised methods make use of sense-annotated corpora, knowledge-based methods exploit the structure and content of the underlying knowledge resource (e.g. definitions or a semantic network).\footnote{Some methods can also be categorized as hybrid, as they make use of both sense-annotated corpora and knowledge resources, e.g., the gloss-augmented model of \citeauthor{Luo2018glosses} \citeyear{Luo2018glosses}.} Currently, supervised methods clearly outperform knowledge-based systems \cite{raganato-camachocollados-navigli:2017:EACLlong}; but, as mentioned earlier, they heavily rely on the availability of sense-annotated corpora, which is generally scarce. 

In this survey we will not go into further details of WSD. For a comprehensive historical overview of WSD we would recommend the survey of \citeauthor{navigli:09} \citeyear{navigli:09}, and a more recent analysis of current methods can be found in the empirical comparison of \citeauthor{raganato-camachocollados-navigli:2017:EACLlong} \citeyear{raganato-camachocollados-navigli:2017:EACLlong}.

\subsection{Notation}
\label{notation}

Throughout this survey we use the following notation. Words will be referred to as $w$ while senses will be written as $s$. Concepts, entities and relations will be referred to as $c$, $e$ and $r$, respectively. Following previous work \cite{navigli:09}, we use the following interpretable expression for senses as well: \textit{word}$_n^p$ is the $n^{th}$ sense of \textit{word} with part of speech $p$. As for synsets as represented in a sense inventory we will use $y$.\footnote{See Section \ref{inventories} for more information about the notions of these resource-related concepts.} A semantic network will be generally represented as $N$.
In order to refer to vectors we will add the vector symbol on the top of each item. For instance, $\vec{w}$ and $\vec{s}$ will refer to the vectors of the word $w$ and sense $s$, respectively. 

In general in this survey we may refer to \textit{sense representation} as a general umbrella term including all vector representations (including embeddings) of meaning beyond the word level, or explicitly to the vector representation of a word associated with a specific meaning\footnote{In some works \textit{senses} have also been referred to as \textit{lexemes} \cite{RotheSchutze:2015}.} (e.g., \textit{bank} with its \textit{financial} meaning), irrespective of whether it comes from a pre-defined sense inventory or not, and whether it refers to a concept (e.g., \textit{banana}) or an entity (e.g., \textit{France}).

\section{Unsupervised Sense Representations}
\label{unsupervised}

Unsupervised sense representations are constructed on the basis of information extracted from text corpora only. 
Word sense induction, i.e., automatic identification of possible meanings of words, lies at the core of these techniques.
An unsupervised model induces different senses of a word by analysing its contextual semantics in a text corpus and represents each sense based on the statistical knowledge derived from the corpus. 
Depending on the type of text corpus used by the model, we can split unsupervised sense representations into two broad categories: (1) techniques that exploit monolingual corpora only (Section \ref{unsuper:mono}) and (2) those exploiting multilingual corpora (Section \ref{unsuper:multi}).

\subsection{Sense Representations Exploiting Monolingual Corpora} 
\label{unsuper:mono}

This section reviews sense representation models that use unlabeled monolingual corpora as their main resource. These approaches can be divided into two main groups: (1) \textbf{clustering-based} (or \textbf{two-stage}) models \cite{Vandecruysetal:2011,ErkPado:2008,Liuetal:2015b}, which first induce senses and then learn representations for these (Section \ref{two-stage}), and (2) \textbf{joint training} \cite{LiJurafsky:2015,Qiuetal:2016}, which perform the induction and representation learning together (Section \ref{jointtraining}). 
Moreover, in Section \ref{contextualized} we will briefly overview \textbf{contextualized embeddings}, an emerging branch of unsupervised techniques which views sense representation from a different perspective.

\subsubsection{Two-Stage Models}
\label{two-stage}

\begin{figure}[t!]
    \includegraphics[width=1\textwidth]{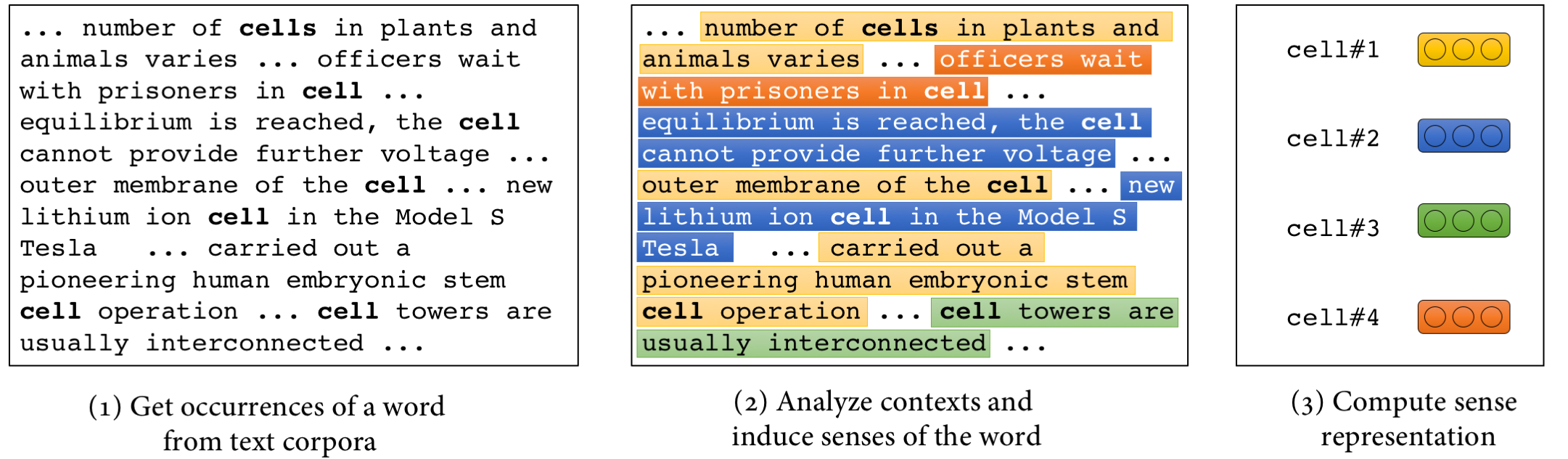}
    \caption{Unsupervised sense representation techniques first induce different senses of a given word (usually by means of clustering occurrences of that word in a text corpus) and then compute representations for each induced sense.}
    \label{fig:unsupervised}
\end{figure}

The \textit{context-group discrimination} of \citeauthor{schutze1998automatic} \citeyear{schutze1998automatic} is one of the pioneering works in sense representation. The approach was an attempt to \textit{automatic} word sense disambiguation in order to address the knowledge-acquisition bottleneck for sense annotated data \cite{Galeetal:92b} and reliance on external resources. The basic idea of context-group discrimination is to automatically induce senses from contextual similarity, computed by \textbf{clustering} the contexts in which an ambiguous word occurs. 
Specifically, each context $C$ of an ambiguous word \textit{w} is represented as a context vector $\vec{v}_C$, computed as the centroid of its content words' vectors $\vec{v}_c$ ($c \in C$).
Context vectors are computed for each word in a given corpus and then clustered into a predetermined number of clusters (context groups) using the Expectation Maximization algorithm \cite[EM]{dempster1977maximum}. 
Context groups for the word are taken as representations for different senses of the word. 
Despite its simplicity, the clustering-based approach of \citeauthor{schutze1998automatic} \citeyear{schutze1998automatic} constitutes the basis for many of the subsequent techniques, which mainly differed in their representation of context or the underlying clustering algorithm.
Figure \ref{fig:unsupervised} depicts the general procedure followed by the two-stage unsupervised sense representation techniques.

Given its requirement for computing independent representations for all individual contexts of a given word, the context-group discrimination approach is not easily scalable to large corpora.
\citeauthor{ReisingerMooney:2010} \citeyear{ReisingerMooney:2010} addressed this by directly clustering the contexts, represented as feature vectors of unigrams, instead of modeling contexts as vectors.
The approach can be considered as the first new-generation sense representation technique, which is often referred to as \textit{multi-prototype}. 
In this specific work, contexts were clustered using Mixtures of von Mises-Fisher distributions (movMF) algorithm.
The algorithm is similar to k-means but permits controlling the semantic breadth using a per-cluster concentration parameter which would better model skewed distributions of cluster sizes. 
%

Similarly, \citeauthor{Huangetal:2012} \citeyear{Huangetal:2012}
proposed a clustering-based sense representation technique with three differences:
(1) context vectors are obtained by a idf-weighted averaging of their word vectors; (2) spherical k-means is used for clustering; and (3) most importantly, occurrences of a word are labeled with their cluster and a second pass is used to learn sense representations.
The idea of two-pass learning has also been employed by \citeauthor{vu2016k} \citeyear{vu2016k} for another sense representation modeling architecture.

     


Sense representations can also be obtained from semantic networks.
For instance, \citeauthor{pelevina2016making} \citeyear{pelevina2016making} 
constructed a semantic graph by connecting each word to the set of its semantically similar words.
Nodes in this network were clustered using the Chinese Whispers algorithm \cite{biemann2006chinese} and senses were induced as a weighted average of words in each cluster.
%
%
A similar sense induction technique was employed by Sense-aware Semantic Analysis \cite[SaSA]{wu2015sense}.
 SaSA follows Explicit Semantic Analysis \cite[ESA]{gabrilovich07} by representing a word using Wikipedia concepts.
Instead of constructing a nearest neighbour graph, a graph of Wikipedia articles is built by gathering all related articles to a word \textit{w} and clustering them. The sense induction step is then performed on the semantic space of Wikipedia articles.


\subsubsection{Joint Models}
\label{jointtraining}

The clustering-based approach to sense representation suffers from the limitation that clustering and sense representation are done independently from each other and, as a result, the two stages do not take advantage from their inherent similarities.
The introduction of embedding models was one of the most revolutionary changes to vector space models of word meaning.
As a closely related field, sense representations did not remain unaffected.
Many researchers have proposed various extensions of the Skip-gram model \cite{Mikolovetal:2013} which would enable the capture of sense-specific distinctions.
A major limitation of the two-stage models is their computational expensiveness\footnote{For instance, the model of \citeauthor{Huangetal:2012} \citeyear{Huangetal:2012} took around one week to learn sense embeddings for a 6,000 subset of the 100,000 vocabulary on a corpus of one billion tokens \cite{Neelakantanetal:2014}.}.
Thanks to the efficiency of embedding algorithms and their unified nature (as opposed to the two-phase nature of more conventional techniques) these techniques are generally efficient.
Hence, many of the recent techniques have relied on embedding models as their base framework.


\citeauthor{Neelakantanetal:2014} \citeyear{Neelakantanetal:2014} was the first to propose a multi-prototype extension of the Skip-gram model.
Their model, called Multiple-Sense Skip-Gram (MSSG), is similar to earlier work in that it represents the context of a word as the centroid of it words' vectors and clusters them to form the target word's sense representation. 
Though, the fundamental difference is that clustering and sense embedding learning are performed jointly.
During training, the intended sense for each word is dynamically selected as the closest sense to the context and weights are updated only for that sense.
In a concurrent work, \citeauthor{tian2014probabilistic} \citeyear{tian2014probabilistic}
proposed a Skip-gram based sense representation technique that significantly reduced the number of parameters with respect to the model of \citeauthor{Huangetal:2012} \citeyear{Huangetal:2012}. 
In this case, word embeddings in the Skip-gram model are replaced with a finite mixture model in which each mixture corresponds to a prototype of the word. The EM algorithm was adopted for the training of this multi-prototype Skip-gram model.

\citeauthor{Liuetal:2015} \citeyear{Liuetal:2015} argued that the above techniques are limited in that they consider only the local context of a word for inducing its sense representations.
To address this limitation, they proposed Topical Word Embeddings (TWE) in which each word is allowed to have different embeddings under different topics, where topics are computed globally using latent topic modelling \cite{blei03latent}. 
Three variants of the model were proposed: (1) TWE-1, which regards each topic as a pseudo word, and learns topic embeddings and word embeddings separately; (2) TWE-2, which considers each word-topic as a pseudo word, and learns topical word embeddings directly; and (3) TWE-3, which assigns distinct embeddings for each word
and each topic and builds the embedding of each word-topic pair by concatenating the corresponding word and topic embeddings. Various extensions of the TWE model have been proposed.
The Neural Tensor Skip-gram (NTSG) model \cite{Liuetal:2015b} applies the same idea of topic modeling for sense representation but introduces a tensor to better learn the interactions between words and topics.
Another extension is MSWE \cite{mixturesenses2017}, which argues that
multiple senses might be triggered for a word in a given context and replaces the selection of the most suitable sense in TWE by a mixture of weights that reflect different association degrees of the word to multiple senses in the context.



These joint unsupervised models, however, suffer from two limitations. First, for ease of implementation, most unsupervised sense representation techniques assume a fixed number of senses per word. This assumption is far from being realistic. Words tend to have a highly variant number of senses, from one (monosemous) to dozens.
In a given sense inventory, usually, most words are monosemous. 
For instance, around 80\% of words in WordNet 3.0 are monosemous, with less than 5\% having more than 3 senses.
However, ambiguous words tend to occur more frequently in a real text which slightly smooths the highly skewed distribution of words across polysemy.
Table \ref{table:sense_distribution} shows the distribution of word types by their number of senses in SemCor \cite{Milleretal:93}, one of the largest available sense-annotated datasets which comprises around 235,000 semantic annotations for thousands of words.
The skewed distribution clearly shows that word types tend to have varying number of senses in a natural text, as also discussed in other studies \cite{piantadosi2014zipf,bennett2016lexsemtm,pasini2018two}.

\begin{table*}[t!]
\setlength{\tabcolsep}{8pt}
\begin{center}
\scalebox{0.9}
{
\begin{tabular}{l cccccccccccc}
\toprule
\bf \# Senses &
2 & 3 & 4 & 5 & 6 & 7 & 8 & 9 & 10 & 11 & 12 & $\ge$ 12\\
\midrule
\bf Nouns & 22\% & 17\% & 14\% & 13\% & 9\% & 7\% & 4\% & 4\% & 3\% & 3\% & 1\% & 3\% \\

\bf Verbs & 15\% & 16\% & 14\% & 13\% & 9\% & 7\% & 5\% & 4\% & 4\% & 3\% & 1\% & 9\% \\

\bf Adjectives & 23\% & 19\% & 15\% & 12\% & 8\% & 5\% & 2\% & 3\% & 3\% & 1\% & 2\% & 6\% \\

\bottomrule

\end{tabular}
}
\end{center}
\caption{\label{table:sense_distribution}
Distribution of words per number of senses in the SemCor dataset (words with frequency $<$ 10 were pruned).}
\end{table*}

Second, a common strand of most unsupervised models is that they extend the Skip-gram model by replacing the conditioning of a word to its context (as in the original model) with an additional conditioning on the intended senses.
However, the context words in these models are not disambiguated.
Hence, a sense embedding is conditioned on the word embeddings of its context. 

In the following we review some of the approaches that are directly targeted at addressing these two limitations of the joint unsupervised models described above:


\begin{enumerate}

\item \textbf{Dynamic polysemy.}
%
A direct solution to the varying polysemy problem of sense representation models would be to set the number of senses of a word as defined by an external sense inventory.
The Skip-gram extension of \citeauthor{nietopina2015} \citeyear{nietopina2015} follows this procedure.
However, by taking external lexicons as groundtruth the approach suffers from two main limitations. First, the model is unable to handle words that are not defined in the lexicon. Second, the model assumes that the sense distinctions defined by the underlying text match those specified by the lexicon, which might not be necessarily true. 
In other words, not all senses of a word might have occurred in the text or the lexicon might not cover all the different intended senses of the word in the underlying text.
A better solution would involve dynamic induction of senses from the underlying text.
Such a model was first implemented in the non-parameteric MSSG (NP-MSSG) system of \citeauthor{Neelakantanetal:2014} \citeyear{Neelakantanetal:2014}.
The model applies the online non-parametric clustering procedure of \citeauthor{Meyerson:2001} \citeyear{Meyerson:2001} to the task by creating a new sense for a word type only if its similarity (as computed using the current context) to existing senses for the word is less than a parameter $\lambda$.
AdaGram \cite{Bartunovetal:2016} improves this dynamic behaviour by a more principled nonparametric Bayesian approach.
The model, which similarly to previous works builds on Skip-gram, assumes that the polysemy of a word is proportional to its frequency (more frequent words are probably more polysemous).

\item \textbf{Pure sense-based models.}
Ideally, a model should model the dependency between sense choices in order to address the ambiguity from context words.
\citeauthor{Qiuetal:2016} \citeyear{Qiuetal:2016} addressed this problem by proposing a pure sense-based model.
The model also expands the disambiguation context from a small window (as done in the previous works) to the whole sentence.
MUSE \cite{lee2017muse} is another Skip-gram extension that proposes pure sense representations using reinforcement learning.
Thanks to a linear-time sense sequence decoding module, the approach provides a more efficient way of searching for sense combinations.

\end{enumerate}

\subsubsection{Contextualized Word Embeddings}
\label{contextualized}

Given that unsupervised sense representations are often produced as a result of clustering, their semantic distinctions are unclear and their mapping to well-defined concepts is not straightforward.
In fact, one of the main limitations of these models lies in their difficult integration into downstream models (more details about this in Section \ref{applications:contextualized}).
Recently, an emerging branch of research has focused on directly integrating unsupervised embeddings into downstream models.
Word embeddings, such as Word2vec and GloVe, compute a single representation for each word, which is used to represent words in downstream models independently from the context in which they appear.
In contrast, contextualized word embeddings are sensitive to the context, i.e., their representation dynamically changes depending on the context in which they appear.

The sequence tagger of \citeauthor{LiMcCallum:2005} \citeyear{LiMcCallum:2005} is one of the pioneering works that employ contextualized representations.
The model infers context sensitive latent variables for each word based on a soft word clustering and integrates them, as additional features, to a CRF sequence tagger.
Since 2011, with the introduction of word embeddings \cite{Collobert:2011,mikolov2013distributed} and the efficacy of neural networks, and in the light of meaning conflation deficiency of word embeddings, context-sensitive models have once again garnered research attention.
Emerging solutions mainly aim at addressing the application limitations of unsupervised techniques; hence, they are generally characterized by their ease of integration into downstream applications. Context2vec \cite{melamud2016context2vec} is one of the earliest and most prominent proposals in the new branch of contextualized representations.
The model represents the context of a target word by extracting the output embedding of a multi-layer perceptron built on top of a bi-directional LSTM language model. Context2vec constitutes the basis for many of the subsequent works.

\begin{figure}[t!]
\centering
    \includegraphics[width=0.9\textwidth]{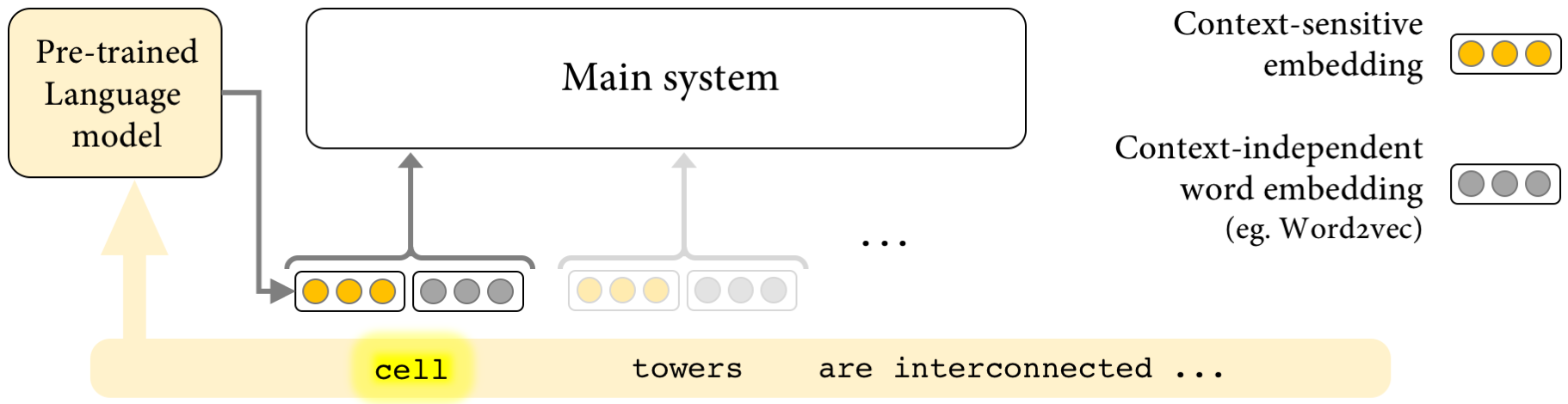}
    \caption{A general illustration of contextualized word embeddings and how they are integrated in NLP models (\textit{Main system} in the figure).
    A language modelling component is responsible for analyzing the context of the target word (\textit{cell} in the figure) and generating its dynamic embedding.
    Unlike (context-independent) word embeddings, which have static representations, contextualized embeddings have dynamic representations that are sensitive to their context. 
    }
    \label{fig:contextualized}
\end{figure}

Figure \ref{fig:contextualized} provides a high-level illustration of the integration of contextualized word embeddings into an NLP model. At the training time, for each word (e.g., \textit{cell} in the figure) in a given input text, the language model unit is responsible for analyzing the context (usually using recurrent neural networks) and adjusting the target word's representation by contextualising (adapting) it to the context.
These context-sensitive embeddings are in fact 
the internal states of a deep recurrent neural network, either in a monolingual language modelling setting \cite{Petersetal:2017,deepcontextual:2018} or a bilingual translation configuration \cite{McCannetal:2017}.
The training of contextualized embeddings is carried out as a pre-training stage, independently from the main task on a large unlabeled or differently-labeled text corpus. 
At the test time, a word's contextualized embeddings is usually concatenated with its static embedding and fed to the main model \cite{deepcontextual:2018}.

The TagLM model of \citeauthor{Petersetal:2017} \citeyear{Petersetal:2017} is a recent example of this branch which trains a multi-layer bidirectional LSTM \cite{HochreiterSchmidhuber:1997} language model on monolingual texts.
The prominent ELMo (Embeddings from Language Models) technique \cite{deepcontextual:2018} is similar in principle with the exception that some weights are shared between the two directions of the language modeling unit.
The Context Vectors (CoVe) model of \citeauthor{McCannetal:2017} \citeyear{McCannetal:2017} similarly computes contextualized representations using a two-layer bidirectional LSTM network, but in the machine translation setting.
CoVe vectors are pre-trained using an LSTM encoder from an attentional sequence-to-sequence machine translation model.\footnote{In general, the pre-training property of contextualized embeddings makes them closely related to transfer learning \cite{Pratt:1992}, which is out of the scope of this article.}

 





\subsection{Sense Representations Exploiting Multilingual Corpora} 
\label{unsuper:multi}

Sense distinctions defined by a sense inventory such as WordNet might not be optimal for some downstream applications, such as Machine Translation (MT).
Given that ambiguity does not necessarily transfer across languages, sense distinctions for MT should ideally be defined based on the translational differences across a specific language pair.
The usual approach to do this is to cluster possible translations of a source word in the target language, with each cluster denoting a specific sense of the source word.

Such translation-specific sense inventories have been used extensively in MT literature \cite{Ideetal:02,carpuat2007improving,Bansal:2012,liuetal:2018handling}.
The same inventory can be used for the creation of sense embeddings that are suitable for MT.
\citeauthor{guo2014learning} \citeyear{guo2014learning} induced a sense inventory in the same manner by clustering words' translations in parallel corpora. Words in the source language were tagged with their corresponding senses and the automatically annotated data was used to compute sense embeddings using standard word embedding techniques.
\citeauthor{ettinger2016retrofitting} \citeyear{ettinger2016retrofitting} followed the same sense induction procedure but used the retrofitting-based sense representation of \citeauthor{jauhar2015ontologically} \citeyear{jauhar2015ontologically}\footnote{See Section \ref{knowledgesense} for more details on this model.}, by replacing the standard sense inventory used in the original model (WordNet) with a translation-specific inventory.

Similarly, \citeauthor{vsuster2016bilingual} \citeyear{vsuster2016bilingual} exploited translation distinctions as supervisory signal in an autoencoder for inducing sense representations.
At the encoding stage, the discrete-state autoencoder assigns a sense to the target word and during decoding recovers the context given the word and its sense.
At training time, the encoder uses words as well as their translations (from aligned corpora). 
This bilingual model was extended by \citeauthor{upadhyay2016beyond} \citeyear{upadhyay2016beyond} to a multilingual setting, in order to better benefit from multilingual distributional information.

\section{Knowledge-Based Semantic Representations}
\label{knowledgebased}

In addition to unsupervised techniques which only learn from text corpora, there is another branch of research which takes advantage of the knowledge available in external resources. This section covers techniques that exploit knowledge resources for constructing sense and concept representations. First, we will give an overview on currently used knowledge resources (Section \ref{inventories}). Then, we will briefly describe some approaches which have made use of knowledge resources for improving word vectors (Section \ref{wordknowledgerep}). Finally, we will focus on the construction of knowledge-based representations of senses (Section \ref{knowledgesense}) and concepts or entities (Section \ref{conceptrepresentations}). 

\subsection{Knowledge Resources}
\label{inventories}

Knowledge resources exist in many flavors. In this section we give an overview of knowledge resources that are mostly used for sense and concept representation learning. The nature of knowledge resources vary with respect to several factors. Knowledge resources can be broadly split into two general categories: expert-made and collaboratively-constructed. Each type has its own advantages and limitations.
Expert-made resources (e.g., WordNet) feature accurate lexicographic information such as textual definitions, examples and semantic relations between concepts. 
On the other hand, collaboratively-constructed resources (e.g., Wikipedia or Wiktionary) provide features such as encyclopedic information, wider coverage, multilinguality and up-to-dateness.\footnote{In addition to these two types of resource, another recent branch is investigating the automatic construction of knowledge resources (particularly WordNet-like) from scratch \cite{khodak2017automated,ustalov2017watset}. However, these output resources are not yet used in practice, and they have been shown to generally lack recall \cite{neale2018lrec}.} 

In the following we describe some of the most important resources in lexical semantics that are used for representation learning, namely WordNet (Section \ref{sec:background:wordnet}), Wikipedia and related efforts (Section \ref{sec:background:wikipedia}), and mergers of different resources such as BabelNet and ConceptNet (Section \ref{sec:background:babelnet}). 


\subsubsection{WordNet}
\label{sec:background:wordnet}

A prominent example of expert-made resource is \textbf{\textsc{WordNet}} \cite{miller1995wordnet}, which is one of the most widely used resources in NLP and semantic representation learning. 
The basic constituents of WordNet are \textit{synsets}.
A synset represents a unique concept which may be expressed through nouns, verbs, adjectives or adverbs and is composed of one or more lexicalizations (i.e., synonyms that are used to express the concept). For example, the synset of the concept defined as ``the series of vertebrae forming the axis of the skeleton and protecting the spinal cord" comprises six lexicalizations: \textit{spinal column}, \textit{vertebral column}, \textit{spine}, \textit{backbone}, \textit{back}, and \textit{rachis}. 
A word can belong to multiple synsets, denoting different meanings it can take.
Hence, WordNet can also be viewd as sense inventory.
The sense definitions in this inventory are widely used in the literature for sense representation learning.

WordNet can alternatively be viewed as a semantic network in which nodes are synsets and edges are lexical or semantic relations  (such as hypernymy or meronymy) which connect different synsets. The most recent version of WordNet version (3.1, released on 2012) covers 155,327 words and 117,979 synsets. In its way to becoming a multilingual resource, WordNet has also been extended to languages other than English through the Open Multilingual WordNet project \cite{bond2013linking} and related efforts.

\subsubsection{Wikipedia, Freebase, Wikidata and DBpedia}
\label{sec:background:wikipedia}

Collaboratively-constructed knowledge resources have had substantial contribution to the research in a wide range of fields, including NLP.
\textbf{\textsc{Wikipedia}} 
is one of the most prominent examples of such resources. Wikipedia is the largest multilingual encyclopedia of world and linguistic knowledge, with individual pages for millions of concepts and entities in over 250 languages. Its coverage is steadily growing, thanks to continuous updates by collaborators. For instance, the English Wikipedia alone receives approximately 750 new articles per day. Each Wikipedia article represents an unambiguous concept (e.g., \textit{Spring (device)}) or entity (e.g., \textit{Washington (state)}), containing a great deal of information in the form of textual information, tables, infoboxes, and various relations such as redirections, disambiguations, and categories. 

A similar collaborative effort was \textbf{Freebase} \cite{bollacker2008freebase}. Partly powered by Wikipedia, Freebase was a large collection of structured data, in the form of a knowledge base. As of January 2014, Freebase contained around over 40 million entities and 2 billion relations. Freebase was finally shut down in May 2016 but its information was partially transferred to Wikidata and served in the construction of Google's Knowledge Graph. \textbf{Wikidata} \cite{wikidata} is a project operated directly by the Wikimedia Foundation with the goal of turning Wikipedia into a fully structured resource, thereby providing a common source of data that can be used by other Wikimedia projects. It is designed as a document-oriented semantic database based on \textit{items}, each representing a topic and identified by a unique identifier. Knowledge is encoded with \textit{statements} in the form of property-value pairs, among which definitions (descriptions) are also included. 
\textbf{DBpedia} \cite{bizer2009dbpedia} is a similar effort towards structuring the content of Wikipedia. In particular, DBpedia exploits Wikipedia infoboxes, which constitutes its main source of information.



\subsubsection{BabelNet and ConceptNet}
\label{sec:background:babelnet}

The types of knowledge available in the expert-based and collaboratively-constructed resources make them often complementary. This has motivated researchers to combine various lexical resources across the two categories \cite{niemann2011people,mccrae2012interchanging,PilehvarNavigli:2014a}. A prominent example is \textbf{\textsc{BabelNet}}
\cite{NavigliPonzetto:12aij}, which provides a merger of WordNet with a number of collaboratively-constructed resources, including Wikipedia. The structure of BabelNet is similar to that of WordNet. Synsets are the main linguistic units and are connected to other semantically related synsets, whose lexicalizations are multilingual in this case. The relations between synsets come from WordNet plus new semantic relations coming from other resources such as Wikipedia hyperlinks and Wikidata. The combination of these resources make BabelNet a large multilingual semantic network, containing 15,780,364 synsets and 277,036,611 lexico-semantic relations for 284 languages in its 4.0 release version.  

\textbf{\textsc{ConceptNet}} \cite{speer2017conceptnet} is a similar resource that combines semantic information from heterogeneous sources. In particular, ConceptNet includes relations from resources like WordNet, Wiktionary and DBpedia, as well as common-sense knowledge from crowdsourcing and games with a purpose. The main difference between ConceptNet and BabelNet lies in their main semantic units: ConceptNet models words whereas BabelNet uses WordNet-style synsets.



\subsection{Knowledge-Enhanced Word Representations}
\label{wordknowledgerep}

As explained in Section \ref{theoretical}, word vector representations (e.g., word embeddings) are mainly constructed by exploiting information from text corpora only. However, there is also a line of research which tries to combine the information available in text corpora with the knowledge encoded in lexical resources. This knowledge can be leveraged to include additional information not available in text corpora in order to improve the semantic coherence or coverage of existing word vector representations. Moreover, knowledge-enhanced word representation techniques are closely related to knowledge-based sense representation learning (see next section), as various models make use of similar techniques interchangeably.

The earlier attempts to improve word embeddings using lexical resources modified the objective function of a neural language model for learning word embeddings (e.g., Skip-gram of Word2vec) in order to integrate relations from lexical resources into the learning process \cite{xu2014rc,yu2014improving}. 
A more recent class of techniques, usually referred to as \textit{retrofitting} \cite{faruqui2014retrofitting}, attempts at improving pre-trained word embeddings with a post-processing step. 
Given any pre-trained word embeddings, the main idea of retrofitting is to move closer words which are connected via a relationship in a given semantic network\footnote{FrameNet \cite{baker1998berkeley}, WordNet and PPDB \cite{ganitkevitch2013ppdb} are used in their experiments.}. The main objective function to minimize in the retrofitting model is the following:

\begin{equation}
\label{retrofitting_equation}
\sum_{i=1}^{|V|} \Big(\alpha_i \|  \vec{w_i}-\vec{\hat{w}}_i\| + \sum_{(w_i,w_j) \in \textsc{N}} \beta_{i,j} \|\vec{w_i}-\vec{w_j}\| \Big)
\end{equation}

\noindent where $|V|$ represents the size of the vocabulary, $\textsc{N}$ is the input semantic network represented as a set of word pairs, $\vec{w_i}$ and $\vec{w_j}$ correspond to word embeddings in the pre-trained model, $\alpha_i$ and $\beta_{i,j}$ are adjustable control values, and $\vec{\hat{w_i}}$ represents the output word embedding.

Building upon retrofitting, \citeauthor{speersemeval2017} \citeyear{speersemeval2017} exploited the multilingual relational information of ConceptNet for constructing embeddings on a multilingual space, and \citeauthor{lengerich2017retrofitting} \citeyear{lengerich2017retrofitting} generalized retrofitting methods by explicitly modeling pairwise relations. 
Other similar approaches are those by \citeauthor{pilehvar-collier:2017:EACLshort} \citeyear{pilehvar-collier:2017:EACLshort} and \citeauthor{goikoetxea2015random} \citeyear{goikoetxea2015random}, which analyze the structure of semantic networks via Personalized Page Rank \cite{Haveliwala:02} for extending the coverage and quality of pre-trained word embeddings, respectively. Finally, \citeauthor{bollegala2016joint} \citeyear{bollegala2016joint} modified the loss function of a given word embedding model to learn vector representations by simultaneously exploiting cues from both co-occurrences and semantic networks.

Recently, a new branch that focuses on specializing word embeddings for specific applications has emerged. 
For instance, \citeauthor{kiela2015specializing} \citeyear{kiela2015specializing} investigated two variants of retrofitting to specialize word embeddings for similarity or relatedness, and \citeauthor{Mrksik:17tacl} \citeyear{Mrksik:17tacl} specialized word embeddings for semantic similarity and dialogue state tracking by exploiting a number of monolingual and cross-lingual linguistic constraints (e.g., synonymy and antonymy) from resources such as PPDB and BabelNet. 

In fact, as shown in this last work, knowledge resources also play an important role in the construction of multilingual vector spaces. The use of external resources avoids the need of compiling a large parallel corpora, which has been traditionally been the main source for learning cross-lingual word embeddings in the literature \cite{upadhyay2016cross,ruder2017survey}.  
These alternative models for learning cross-lingual embeddings exploit knowledge from lexical resources such as WordNet or BabelNet \cite{Mrksik:17tacl,goikoetxea2018bilingual}, bilingual dictionaries \cite{mikolov2013exploiting,ammar2016massively,artetxe2016learning,doval:meemiemnlp2018} or comparable corpora extracted from Wikipedia \cite{vulic2015bilingual}.

\subsection{Knowledge-Based Sense Representations}
\label{knowledgesense}

\begin{figure}[t!]
    \includegraphics[width=1\textwidth]{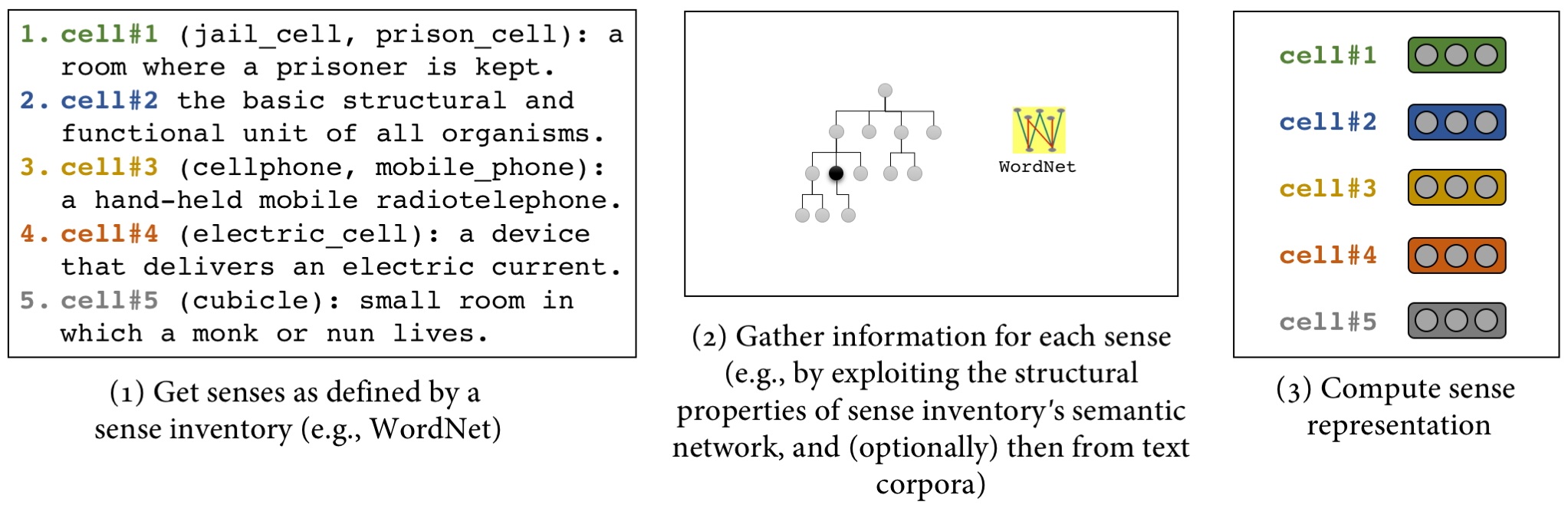}
    \caption{Knowledge-based sense representation techniques take sense distinctions for a word as defined by an external lexical resource (sense inventory). For each sense, relevant information is gathered and a representation is computed.}
    \label{fig:knowledge-based}
\end{figure}

This section provides an overview of the state of the art in knowledge-based sense representations. These representations are usually obtained as a result of \textit{de-conflating} a word into its individual sense representations, as defined by an external sense inventory.
Figure \ref{fig:knowledge-based} depicts the main workflow for knowledge-based sense vector representation modeling techniques. 
The learning signal for these techniques vary, but in the main two different types of information available in lexical resources are leveraged: textual definitions (or \textit{glosses}) and semantic networks.

\textbf{Textual definitions} are used as main signals for initializing sense embeddings by several approaches. \citeauthor{chenunified:2014} \citeyear{chenunified:2014} proposed an initialization of word sense embeddings by averaging pre-trained word embeddings trained on text corpora. Then, these initialized sense representations are utilized to disambiguate a large corpus. Finally, the training objective of Skip-gram from Word2vec \cite{Mikolovetal:2013} is modified in order to learn both word and sense embeddings from the disambiguated corpus. In contrast, \citeauthor{chen2015improving} \citeyear{chen2015improving} exploited a convolutional neural network architecture for initializing sense embeddings using textual definitions from lexical resources. Then, these initialized sense embeddings are fed into a variant of the Multi-sense Skip-gram Model of \citeauthor{Neelakantanetal:2014} \citeyear{Neelakantanetal:2014} (see Section \ref{unsuper:mono}) for learning knowledge-based sense embeddings. Finally, in \citeauthor{Yang2016291} \citeyear{Yang2016291} word sense embeddings are learned by exploiting an adapted Lesk\footnote{The original Lesk algorithm \cite{Lesk:86} and its variants exploit the similarity between textual definitions and a target word's context for disambiguation.} algorithm \cite{vasilescu2004evaluating} over short contexts of word pairs.

A different line of research has experimented with the graph structure of lexical resources for learning knowledge-based sense representations. As explained in Section \ref{inventories}, many of the existing lexical resources can be viewed as \textbf{semantic networks} in which nodes are concepts and edges represent the relations among concepts. Semantic networks constitute suitable knowledge resources for disambiguating large amounts of text \cite{agirre2014random,Moroetal:14tacl}. Therefore, a straightforward method to learn sense representations would be to automatically disambiguate text corpora and apply a word representation learning method on the resulting sense-annotated text \cite{iacobacci:2015}. Following this direction, \citeauthor{mancini2017sw2v} \citeyear{mancini2017sw2v} proposed a shallow graph-based disambiguation procedure and modified the objective functions of Word2vec in order to simultaneously learn word and sense embeddings in a shared vector space. 
The objective function is in essence similar to the objective function proposed by \citeauthor{chenunified:2014} \citeyear{chenunified:2014} explained before, which also learns both word and sense embeddings in the last step of the learning process.

Similarly to the post-processing of word embeddings by using knowledge resources (see Section \ref{wordknowledgerep}), recent works have made use of pre-trained word embeddings not only for improving them but also de-conflating them into senses. Approaches that \textbf{post-process pre-trained word embeddings} for learning sense embeddings are listed below:

\begin{enumerate}

    \item  One way to obtain sense representations from a semantic network is to directly apply the Personalized PageRank algorithm \cite{Haveliwala:02}, as done by \citeauthor{PilehvarNavigli:2015aij} \citeyear{PilehvarNavigli:2015aij}. The algorithm carries out a set of random graph walks to compute a vector representation for each WordNet synset (node in the network). 
    Using a similar random walk-based procedure, \citeauthor{PilehvarCollier:2016emnlp} \citeyear{PilehvarCollier:2016emnlp} extracted for each WordNet word sense a set of \textit{sense biasing words}.
    Based on these, they
    put forward an approach, called DeConf, which takes a pre-trained word ebmeddings space as input and adds a set of sense embeddings (as defined by WordNet) to the same space. DeConf achieves this by pushing a word's embedding in the space to the region occupied by its corresponding sense biasing words (for a specific sense of the word).
    Figure \ref{fig:deconf} shows the word \textit{digit} and its induced \textit{hand} and \textit{number} senses in the vector space. 

\begin{figure}[t!]
    \centering
    \includegraphics[width=0.5\textwidth]{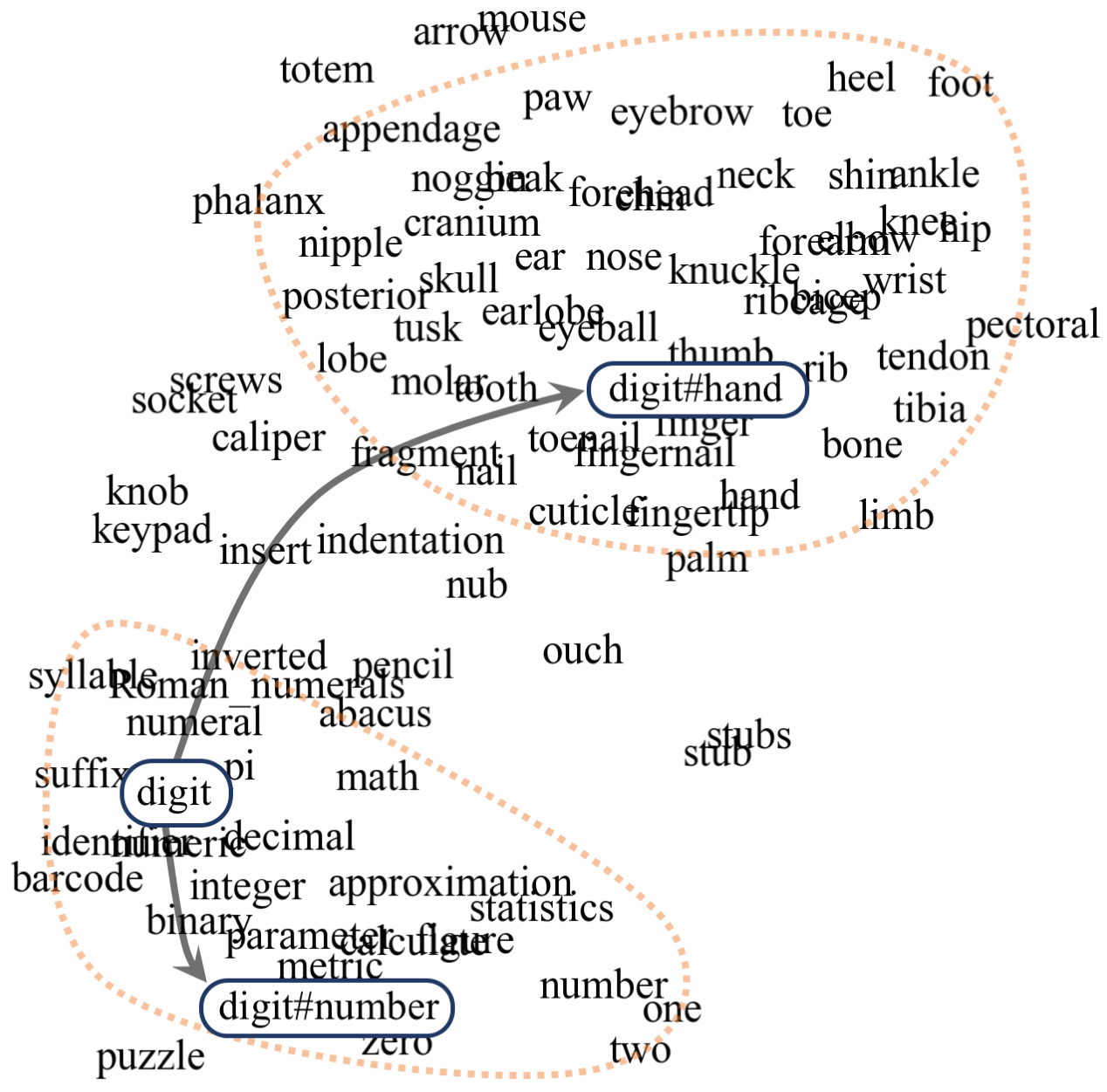}
    \caption{A mixed semantic space of words and word senses. DeConf \cite{PilehvarCollier:2016emnlp} introduces two new points in the word embedding space, for the \textit{mathematical} and \textit{body part} senses of the word \textit{digit}, resulting in the mixed space.
    }
    \label{fig:deconf}
\end{figure}

    \item \citeauthor{jauhar2015ontologically} \citeyear{jauhar2015ontologically} proposed an extension of \textit{retrofitting}\footnote{See Section \ref{wordknowledgerep} for more information on retrofitting.} \cite{faruqui2014retrofitting} for learning representations for the senses of the underlying sense inventory (e.g., WordNet). They additionally presented a second approach which adapts the training objective of Word2vec to include senses within the learning process. The training objective is optimized using EM.
    
    \item  \citeauthor{johansson2015embedding} \citeyear{johansson2015embedding} post-processed pre-trained word embeddings through an optimization formulation with two main constraints: polysemous word embeddings can be decomposed as combinations of their corresponding sense embeddings and sense embeddings should be close to their neighbours in the semantic network. A Swedish semantic network, SALDO \cite{borin2013saldo}, was used in their experiments, although their approach may be directly extensible to different semantic networks as well. 
    
    \item  Finally, AutoExtend \cite{RotheSchutze:2015} is another method using pre-trained word embeddings as input. In this case, they put forward an autoencoder architecture based on two main constraints: a word vector corresponds to the sum of its sense vectors and a synset to the sum of its lexicalizations (senses). For example, the vector of the word \textit{crane} would correspond to the sum of the vectors for its senses \textit{crane}$_n^1$, \textit{crane}$_n^2$ and \textit{crane}$_v^1$ (using WordNet as reference). Similarly, the vector of the synset defined as ``arrange for and reserve (something for someone else) in advance" in WordNet would be equal to the sum of the vectors of its corresponding senses \textit{reserve}, \textit{hold} and \textit{book}. Equation \ref{autoextend_equation} displays these constraints mathematically:

\begin{equation}
\label{autoextend_equation}
\vec{w}=\sum_{i=1}^{n} \vec{s_i}\, ;\ \vec{y}=\sum_{j=1}^{m} \vec{s_j},
\end{equation}
\noindent where $s_i$ and $s_j$ refer to the senses of word $w$ and synset $y$, respectively.

\end{enumerate}







\subsection{Concept and Entity Representations}
\label{conceptrepresentations}


In this section we present approaches which rely solely on the relational information of knowledge bases (Section \ref{knowledgebasemebeddings}) and hybrid models which combine cues from text corpora and knowledge resources (Section \ref{hybrid}).

\subsubsection{Knowledge Base Embeddings}
\label{knowledgebasemebeddings}

This section provides a review of those representation techniques targeting concepts and named entities from knowledge bases only. A large body of research in this area takes knowledge graphs (or semantic networks) as signals to construct representations of entities (and relations), specifically targeted to the knowledge base completion task\footnote{Given an incomplete knowledge base as input, the knowledge base completion task consists of predicting relations which were missing in the original resource.}.

\begin{figure}[t!]
    \centering
    \includegraphics[width=0.8\textwidth]{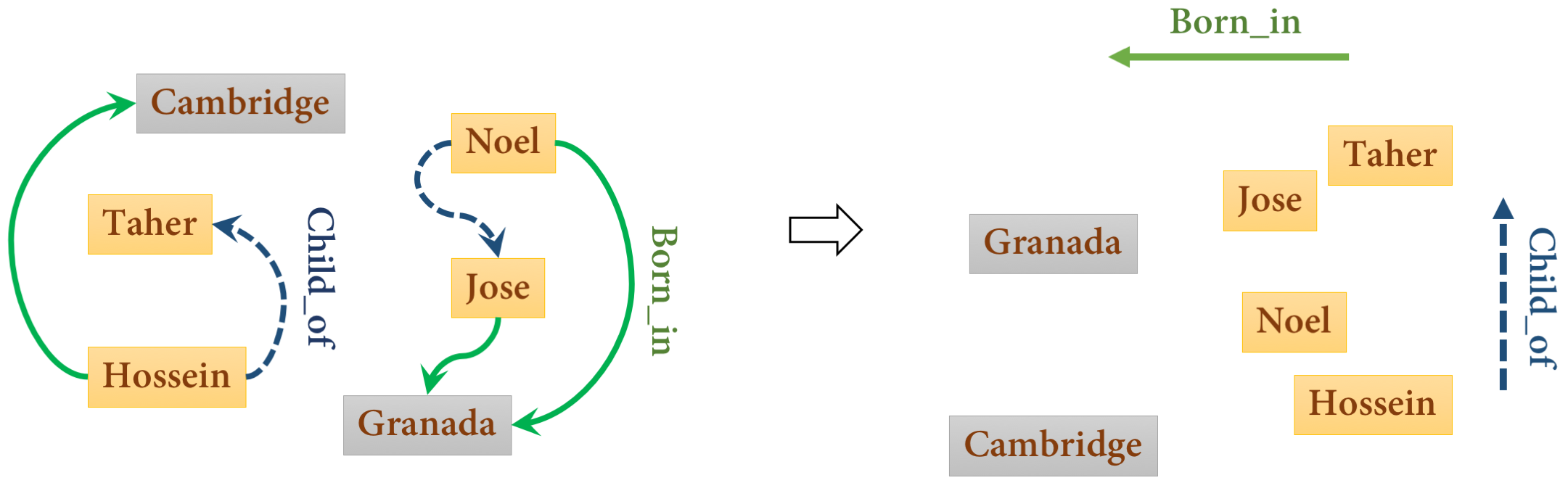}
    \caption{From a knowledge graph to entity and relation embeddings. Illustration idea is based on the slides of \citeauthor{weston2014embedding} \citeyear{weston2014embedding}.}
    \label{fig:transe}
\end{figure}

A pioneering work in this area is TransE \cite{bordes2013translating}, a method to embed both entities and relations. In this model relations are viewed as translations which operate in the same vector space as entities. Given a knowledge base represented as a set of triples $\{(e_1,r,e_2)\})$, where $e_1$ and $e_2$ are entities and $r$ the relation between them, the main goal is to approach the entities in a way that $\vec{e_1}+\vec{r} \approx \vec{e_2}$ for all triples in the space (i.e., $\forall (e_1,r,e_2) \in N$). Figure \ref{fig:transe} illustrates the main idea behind the model. This objective may be achieved by exploiting different learning architectures and constraints. In the original work of \citeauthor{bordes2013translating} \citeyear{bordes2013translating}, the optimization is carried out by stochastic gradient descent with an $L_2$ normalization of embeddings as an additional constraint. Following this underlying idea, various approaches have proposed improvements of different parts of the learning architecture:

\begin{enumerate}

\item TransP \cite{wang2014knowledgetranslating} is a similar model that provides improvements on the relational mapping by dealing with specific properties present in the knowledge graph.

\item \citeauthor{lin2015learning} \citeyear{lin2015learning} proposed to learn embeddings of entities and relations in separate spaces (TransR).

\item \citeauthor{ji2015knowledge} \citeyear{ji2015knowledge} introduced a dynamic mapping for each entity-relation pair in separated spaces (TransD).

\item \citeauthor{luo2015context} \citeyear{luo2015context} put forward a two-stage architecture using pre-trained word embeddings for initialization.

\item A unified learning framework that generalize TransE and NTN \cite{SocherEtAl2013:RNTN} was presented by \citeauthor{yang2014embedding} \citeyear{yang2014embedding}.

\item Finally, \citeauthor{ebisu2017toruse} \citeyear{ebisu2017toruse} discussed regularization issues from TransE and proposed TorusE, which benefits from a new regularization method solving TransE's regularization problems. 

\end{enumerate}

Alternatively, a branch of research focuses specifically on modeling entities only (not relations) and computes embeddings for individual nodes in the graph. DeepWalk \cite{Perozzi:2014} is one of the prominent techniques in this branch. The core idea in this algorithm is to use random graph walks to represent a given graph as a series of artificial sentences. Similarly to natural language in which semantically similar words tend to co-occur, consecutive words in these artificial sentences correspond to neighbouring (topologically related) vertices in the graph. These sentences are then used as input to the Skip-gram model (see Section \ref{wordembeddings}) and embeddings for individual words (i.e., concept nodes) are computed. Node2vec \cite{Grover:2016} is an extension of DeepWalk which better controls the depth-first and breadth-first property of random walks. In contrast, \citeauthor{poincare_NIPS2017} \citeyear{poincare_NIPS2017} put forward a newer form of representation by embedding words into a Poincar\'{e} ball\footnote{A Poincar\'{e} ball is a hyperbolic space in which all points are inside the unit disk.} which takes into account both similarity and the hierarchical structure of the taxonomy given as input\footnote{WordNet is used as the reference taxonomy in the original work.}. 

These have been some of the most relevant works on knowledge base embeddings in recent years, but given the multitude of papers on this topic, this review was by no means comprehensive. A broader overview of knowledge graph embeddings, including more in-depth explanations, 
is presented by \citeauthor{cai2018comprehensive} \citeyear{cai2018comprehensive} or \citeauthor{nguyen2017overview} \citeyear{nguyen2017overview}, the latter focusing on the knowledge base completion task.      



\subsubsection{Hybrid Models Exploiting Knowledge Bases and Text Corpora}
\label{hybrid}

In addition to techniques that entirely rely on the information available in knowledge bases, there are models that combine cues from both knowledge bases and text corpora into the same representation. Given its semi-structured nature and the textual content provided, Wikipedia has been the main source for these kind of representations. While most approaches make use of Wikipedia-annotated corpora as their main source to learn representations for Wikipedia concepts and entities \cite{wang2014knowledge,sherkat2017vector,cao2017bridge}, the combination of knowledge from heterogeneous resources like Wikipedia and WordNet has also been explored \cite{camacho2016nasari}.\footnote{The combination of Wikipedia and WordNet relies on the multilingual mapping provided by BabelNet (see Section \ref{sec:background:babelnet} for more information about BabelNet).}  

Given their hybrid nature, these models can easily be used in textual applications as well. A straightforward application is word or named entity disambiguation, for which the embeddings can be used as initialization in the embedding layer on a neural network architecture \cite{fang2016entity,eshel2017named} or used directly as a knowledge-based disambiguation system exploiting semantic similarity \cite{camacho2016nasari}.







\section{Evaluation}
\label{sec:evaluation}

In this section we present the most common evaluation benchmarks for assessing the quality of meaning representations.  Depending on their nature, evaluation procedures are generally divided into intrinsic (Section \ref{similarity}) and extrinsic (Section \ref{extrinsicevaluation}). 




\subsection{Intrinsic Evaluation}
\label{similarity}

Intrinsic evaluation refers to a class of benchmarks that provide a generic evaluation of the quality and coherence of a vector space, independently from their performance in downstream applications. Different properties can be intrinsically tested, with semantic similarity being traditionally viewed as the most straightforward feature to evaluate meaning representations. In particular, the semantic similarity of small lexical units such as words and senses, in which compositionality is not required, has received the most attention. Word similarity datasets exist in many flavors. It is also worth distinguishing the notions of similarity and relatedness. While words that are semantically similar can be technically substituted with each other in a context, related words are enough to co-occur in the same context (e.g., within a document) without the need for substitutability. WordSim-353 \cite{Levetal:2002} is a dataset that conflates these two notions. Genuine similarity datasets include RG-65 \cite{RG65:1965}, which only contains 65 word pairs, or SimLex-999 \cite{hill2015simlex}, consisting of 999 word pairs. Moreover, there are multilingual benchmarks which include word similarity datasets for several languages. For instance, the translations and reannotations of WordSim-353 and SimLex-999 \cite{leviant2015separated} and the datasets from the SemEval-2017 task on multilingual word similarity \cite{semeval2017similarity} provide evaluation benchmarks for languages other than English.

In order to adapt these word-based evaluation benchmarks to sense vectors, various strategies have been proposed \cite{ReisingerMooney:2010}. Among these, the most popular is to take the most similar pair of senses across the two words \cite{Resnik:95,PilehvarNavigli:2015aij,mancini2017sw2v}, also known as \textit{MaxSim}:

\begin{equation}
sim(w_1,w_2)= \max_{s_1 \in {S_{w_1}}, s_2 \in S_{w_2}} \cos(\vec{s}_1,\vec{s}_2)
\end{equation}
%

\noindent where $S_{w_i}$ is a set including all senses of $w_i$ and $\vec{s}_i$ represents the sense vector representation of the sense $s_i$. Another strategy, known as \textit{AvgSim}, simply averages the pairwise similarities of all possible senses of $w_1$ and $w_2$. Cosine similarity ($cos$) is the most prominent metric for computing the similarity between sense vectors.

In all these benchmarks, words are paired in isolation. However, we know that for a specific meaning of an ambiguous word to be triggered, the word needs to appear in particular contexts. In fact, \citeauthor{Kilgarriff:97b} \citeyear{Kilgarriff:97b} argued that representing a word with a fixed set of senses may not be the best way for modelling word senses but instead, word senses should be defined according to a given context. To this end, \citeauthor{Huangetal:2012} \citeyear{Huangetal:2012} presented a different kind of similarity dataset in which words are provided with their corresponding contexts.
The task consists of assessing the similarity of two words by taking into consideration the contexts in which they occur. 
The dataset is known as Stanford Contextual Word Similarity (SCWS) and has been established as one of the main intrinsic evaluations for sense representations. 
A pre-disambiguation step is required to leverage sense representations in this task. Simple similarity measures such as \textit{MaxSimC} or \textit{AvgSimC} are generally utilized. Unlike \textit{MaxSim} and \textit{AvgSim}, \textit{MaxSimC} and \textit{AvgSimC} take the context of the target word into account. 
First, the confidence for selecting the most appropriate sense within the sentence is computed (e.g., by computing the average of word embeddings from the context and selecting the sense which is closest to the average context vector in terms of cosine similarity). Then, the final score corresponds to the similarity between the selected senses (i.e., \textit{MaxSimC}) or to a weighted average among all senses (i.e., \textit{AvgSimC}). 

However, even though sense representations have generally outperformed word-based models on this dataset, the simple strategies used to disambiguate the input text may not have been optimal. In fact, it has been recently shown that the improvements of sense-based models in word similarity tasks using \textit{AvgSim} may not be due to accurate meaning modeling but to related artifacts such as sub-sampling, which had not been controlled for \cite{dubossarskycoming}. This goes in line with a recent study analyzing how well sense and contextualized representations capture meaning in context \cite{pilehvar2018wic}. 
The binary classification task proposed in this analysis consists of deciding whether the occurrences of a target word in two different contexts correspond to the same meaning or not. The results showed how recent sense\footnote{Similarly to the \textit{MaxSimC} technique, sense representations were evaluated by retrieving the closest sense embedding to the context-based vector, computed by averaging its word embeddings.} and contextualized representation techniques fail at accurately distinguishing meanings in context, performing only slightly better than a simple baseline, while significantly lagging behind the human inter-rater agreement of the dataset.\footnote{Another study which took the hypernymy detection task as test bed for their experiments \cite{vyas2017detecting} came to similar conclusions.} 


Finally, in addition to these tasks, there exist other intrinsic evaluation procedures such as synonymy selection \cite{Landauer:1997,Turney:2001,JarmaszSzpakowicz:2003,ReisingerMooney:2010}, outlier detection \cite{camacho2016find,blair2016automated,Stanovsky2018oddman} or sense clustering \cite{Snowetal:2007,Dandalaetal:2013}. For more information, \citeauthor{bakarov2018survey} \citeyear{bakarov2018survey} provides a more comprehensive overview of intrinsic evaluation benchmarks.




\subsection{Extrinsic Evaluation}
\label{extrinsicevaluation}



Extrinsic evaluation procedures aim at assessing the quality of meaning representations within a downstream task. In addition to intrinsic evaluation procedures, extrinsic evaluation is necessary to understand the effectiveness of different sense representation techniques in real-world applications. This is especially relevant because intrinsic evaluation protocols do not always correlate with downstream performance \cite{tsvetkov2015evaluation,Chiu-repeval:16,faruqui2016repeval}. 
However, while extrinsic evaluation is definitely important to assess the effectiveness of integrating sense representations in downstream tasks, there is also a higher variability in terms of tasks, pipelines and benchmarks in comparison to intrinsic procedures, which are more straightforward. 

Some of the most common tasks that have been used as extrinsic evaluation procedures for sense representations in natural language processing are text categorization and sentiment analysis \cite{Liuetal:2015,LiJurafsky:2015,pilehvaracl17}, document similarity \cite{wu2015sense}, and word sense induction \cite{pelevina2016making,panchenko-EtAl:2017:EACLlong} and disambiguation \cite{chenunified:2014,RotheSchutze:2015,camacho2016nasari,deepcontextual:2018}. 
As mentioned in Section \ref{knowledgebasemebeddings}, knowledge base embeddings are also frequently evaluated on the knowledge base completion task \cite{bordes2013translating}. In the following section we will explain in more detail some of the applications to which sense representations have been applied to date.

\section{Applications}
\label{applications}


As mentioned in Section \ref{sec:evaluation} and throughout the survey, one of the main goals of research in meaning representations is to enable effective integration of these knowledge carriers into downstream applications. Unlike word representations (and more specifically embeddings), sense representations are still in their infancy in this regard. This is also due to the non-immediate integration of these representations, which generally requires an additional word sense disambiguation or induction step. However, as with word embeddings, sense representations can be theoretically applied to multiple applications. 


The integration of sense representations into downstream applications is not a new trend. Since the nineties, many heterogeneous efforts have emerged in this direction for important text-based applications, with varying degree of success. \textbf{Information retrieval} has been one of the first applications in which the integration of word senses was investigated. In one of the earlier attempts, \citeauthor{SchutzePedersen:95} \citeyear{SchutzePedersen:95} showed how document-query similarity based on word senses could lead to considerable improvements with respect to word-based models. 

Another classic task which has witnessed recurring efforts to incorporate sense-level information is \textbf{Machine Translation} (MT). Since a word may have different translations depending on its intended meaning in a context, sense identification has been traditionally believed to be able to potentially improve word-based MT models. Carpuat et al. analyzed the impact of WSD in the performance of standard MT systems at the time \cite{carpuat2005word,carpuat2007phrase,carpuat2007improving}. The studies were inconclusive, but generally reflected the difficulty to successfully integrate semantically-grounded models into an MT pipeline. This was also partially due to the lack of sense-annotated corpora, producing the knowledge-acquisition bottleneck \cite{Galeetal:92b}.\footnote{It is worth mentioning that, while there is still a lack of sense-annotated multilingual corpora, recent efforts have directly addressed this issue by (semi-)automatically disambiguating large amounts of parallel corpora \cite{taghipour2015one,otegi:2016,dellibovietal:17}.} 

Since then, word senses (and in particular sense representations) have been integrated into various NLP tasks. In the following we discuss the application of sense representations (Section \ref{applications:sense}) and the more recent contextualized representations (Section \ref{applications:contextualized}) to downstream tasks.

\subsection{Application of sense representations}
\label{applications:sense}

The integration of unsupervised sense representations into downstream applications is limited in the literature. 
\citeauthor{LiJurafsky:2015} \citeyear{LiJurafsky:2015} proposed a framework to integrate unsupervised sense embeddings into various natural language processing tasks. 
The research concluded that the proposed unsupervised representations did not provide a significant influence, suggesting that an increase in the dimensionality of word embeddings can lead to similar results. However, the disambiguation step was a simple procedure based on the similarity between sense embeddings and an embedding representation of the input text (computed as the average of the content words' embeddings). A more recent proposal is the Multi-Sense LSTM model of \citeauthor{kartsaklisetal:2018} \citeyear{kartsaklisetal:2018}, which avoids the need for an explicit disambiguation. The system replaces in a neural network the conventional word embedding layer for each word with $k$ separate sense embeddings. For each training instance, the intended sense is dynamically selected using an attention mechanism and correspondingly updated. A similar mechanism is used at test time based on the context in which the word appears. This system proved effective in various tasks, reporting state-of-the-art performance across multiple benchmarks in text-to-entity mapping. 


As far as knowledge-based representations are concerned, an explicit or implicit word sense disambiguation step is required to transform words into their intended senses.
\citeauthor{pilehvaracl17} \citeyear{pilehvaracl17} proposed a method based on a shared space of word and knowledge-based sense embeddings, introducing a simple graph-based disambiguation step prior to their integration into a neural network architecture for text classification. The inclusion of senses was shown to improve when the input text was large enough, but the inclusion of pre-trained sense embeddings in this setting did not significantly improve the use of word embeddings in most datasets. The major benefits of using sense representations were observed when using supersenses (see Section \ref{granularity}), a conclusion which was also observed by \citeauthor{flekovasupersense} \citeyear{flekovasupersense} on other downstream classification tasks.



In addition to these studies, 
there have been other applications in which sense and concept representations, in their broader meaning, have been effectively integrated: 
word sense or named entity disambiguation \cite{chenunified:2014,RotheSchutze:2015,camacho2016nasari,fang2016entity,panchenko2017using,deepcontextual:2018}, knowledge base completion \cite{bordes2013translating} or unification \cite{delli2015knowledge}, common-sense reasoning \cite{lieto2017towards}, lexical substitution \cite{cocos2016word}, hypernym discovery \cite{EspinosaEMNLP2016}, lexical entailment \cite{poincare_NIPS2017},  
or visual object discovery \cite{young2017semantic}.

\subsection{Application of contextualized representations}
\label{applications:contextualized}

Contextualized representations (see Section \ref{contextualized}) offer an alternative solution to the problem of having to discretize the input sentence into word senses, by employing a different strategy. 
In this case, the input words are not explicitly replaced with sense embeddings; however, their representations are dynamically adjusted according to the context (hence, an implicit disambiguation).

Thanks to their dynamic nature, contextualized word embeddings feature seamless integration into neural architectures and, 
 therefore, they have been evaluated in a wide range of NLP tasks, including sentiment analysis, question answering and classification, textual entailment, semantic role labeling, reading comprehension, named entity extraction, and coreference resolution \cite{deepcontextual:2018,salant-berant:2018,McCannetal:2017}.
Improvements have been reported upon substituting conventional static word embeddings with their contextualized counterparts, proving the advantage of having a dynamic representation that can adapt the semantics of a target word based on its context.
Other recent examples include the HIT-SCIR system of \citeauthor{conll-winner:2018} \citeyear{conll-winner:2018}, which attained the best performance in the CoNLL 2018 shared task on universal dependency parsing \cite{zeman-EtAl:2018:K18-2} by employing ELMo embeddings \cite{deepcontextual:2018}, and  the end-to-end neural machine translation architecture of \citeauthor{liuetal:2018handling}  \citeyear{liuetal:2018handling}, which explicitly models homographs (i.e., ambiguous words) with context-aware embeddings, achieving improved translation performance for ambiguous words.

\section{Analysis}
\label{analysis}

This section provides an analysis and comparison of knowledge-based and unsupervised representation techniques, highlighting the advantages and limitations of each, while suggesting the settings and scenarios for which each technique is suited. We focus on four important aspects: interpretability (Section \ref{interpretability}), adaptability to different domains (Section \ref{adaptability}), sense granularity (Section \ref{granularity}), and compositionality (Section \ref{compositionality}).  

\subsection{Interpretability}
\label{interpretability}

One of the main reasons behind moving from word to sense level is the semantically-grounded nature of word senses, which may enable a better interpretability. 
In this particular aspect, however, there is a considerable difference between unsupervised and knowledge-based models. 
Unsupervised models learn senses directly from text corpora, which results in model-specific sense interpretations. 
These induced senses do not necessarily correspond to human notions of sense distinctions, or are not easily distinguishable. 
For this reason, methods have been proposed to improve the interpretability of unsupervised sense representations, either by extracting their hypernyms or their visual representations (i.e., an image illustrating a specific meaning) \cite{panchenko-EtAl:2017:EACLlong} or by mapping the induced senses to external sense inventories \cite{panchenko2016best}.

In contrast, knowledge-based representations are already linked to entries in a sense inventory, which enables a higher interpretability, as these entries are generally associated with definitions, examples, images and often relations with other concepts (e.g., WordNet) and translations (e.g., BabelNet). This, in turn, enables the direct injection of extra prior information from lexical resources, which may be useful to supply end models with a deeper background knowledge \cite{young2017semantic}. As a drawback, knowledge-based representations are generally constrained to the underlying sense inventories and, hence, may fail to provide an accurate representation of unseen novel senses in text corpora. This is partially solved by keeping sense inventories updated, though not generally a straightforward process. As explained in Section \ref{inventories}, collaborative resources like Wikipedia are less prone by this issue.

\subsection{Adaptability to Different Domains}
\label{adaptability}

One feature which has been praised in word embeddings is their adaptability to general and specialized domains \cite{goldberg2016primer}. From this aspect, unsupervised models have a theoretical advantage over knowledge-based counterparts as they are able to directly induce senses from a given text corpus. This provides them with the chance to adapt their sense distinctions according to the domain at hand and to the given task.
In the contrary, knowledge-based systems generally learn representations for all senses given by a sense inventory; hence, they are unable to specialize their sense distinctions to the domain or adapt their granularity to the task.

Knowledge-enhanced approaches like those proposed by \citeauthor{mancini2017sw2v} \citeyear{mancini2017sw2v} or \citeauthor{fang2016entity} \citeyear{fang2016entity}, which directly learn from text corpora, may partially alleviate this limitation of knowledge-based models. However, the senses should still be present in the semantic network used as input for the model. In other words, knowledge-based approaches are not able to learn new senses, which may be an important limitation in some specific domains and tasks. Moreover, the accurate representation of certain domains would require suitable knowledge resources, which might not be available for specialized domains or low-resource languages.

\subsection{Sense Granularity}
\label{granularity}

A sense inventory may list a few dozen different senses for words such as \textit{run}, \textit{play} and \textit{get}.  
Words with multiple senses (i.e., ambiguous) are generally classified into two categories: polysems and homonyms. Polysemous words have multiple related meanings. 
For instance the word \textit{mark} can refer to a ``distinguishing symbol" as well as a ``visible indication made on a surface". In this case the distinctions of these two senses are also said to be fine-grained, as these two meanings are difficult to be torn apart. Homonymous words\footnote{According to the Cambridge Dictionary, a homonym is ``a word that sounds the same (homophone) or is spelled the same (homograph) as another word but has a different meaning''. Given that NLP focuses on written forms, a homonym in this context usually refers to the latter condition, i.e., homographs with different meanings.} have meanings that are completely unrelated. For instance, the geological and financial institution senses of the word \textit{bank}\footnote{The distinction between homonyms and polysems can sometime be subtle. For instance, research in historical linguistics has shown that the two meanings of the word \textit{bank} could have been related to each other earlier in the Italian language, since the bankers used to do their business on the riverbanks.}. This would also be a case of a coarse-grained distinction of senses, as these two meaning of \textit{bank} are clearly different.

In general, the fine granularity of some sense inventories has always been a point of argument in NLP \cite{Kilgarriff:97b,navigli:09,Hovyetal:13}. It has been pointed out that sense distinctions in WordNet might be too fine-grained to be useful for many NLP applications \cite{Navigli:06b,Snowetal:2007,Hovyetal:13}. For instance, WordNet 3.0 (see Section \ref{sec:background:wordnet}) lists 41 different senses for the verb \textit{run}. However, most of these senses are translated to either \textit{correr} or \textit{operar} in Spanish. Therefore, a multilingual task such as machine translation might not benefit from the additional distinctions provided by the sense inventory. In fact, a merging of these fine-grained distinctions into more coarse-grained classes (referred to as \textit{supersenses} in WordNet) has been shown to be beneficial in various downstream applications \cite{flekovasupersense,pilehvaracl17}.

This discussion is also relevant for unsupervised techniques. The dynamic learning of senses, instead of fixing the number of senses for all words, has shown to provide a more realistic distribution of senses (see Section \ref{jointtraining}). 
Moreover, there have been discussions about whether all occurrences of words can be effectively partitioned into senses \cite{Kilgarriff:97b,Hanks:00,kilgarriff2007word}, leading to a new scheme in which meanings of a word are described in a graded fashion \cite{erk2009investigations,mccarthy2016word}. While the scheme is not covered in this survey, it has been shown that a graded scale to assess senses may correlate better to how humans perceive different meanings. Although not exactly the same conclusions, these findings are also related to the criticisms about the fine granularity of current sense inventories, which has shown to be harmful in certain downstream applications.

\subsection{Compositionality}
\label{compositionality}

Compositional methods model the semantics of a complex expression based on the meanings of its constituents (e.g., words).  
Typically, constituent words are represented as their word vector with all the meanings conflated.
However, for an ambiguous word in an expression, usually only a single meaning is triggered and other senses are irrelevant.
Therefore, pinpointing the meaning of a word to the given context may be a reasonable idea for compositionality.
This can be crucial to applications such as information retrieval in which query ambiguity can be an issue \cite{AllanRaghavan:02,di2013clustering}.

Different works have tried to introduce sense representations in the context of compositionality \cite{koper2017applying,kober-EtAl:2017:SENSE2017}, with different degrees of success. The main idea is to select the intended sense of a word and only introduce that specific meaning into the composition, either through context-based sense induction \cite{Thateretal:2011}, exemplar-based representation \cite{Reddyetal:2011}, or with the help of external resources, such as WordNet \cite{gamallo-pereirafarina:2017:SENSE2017}. 
An example of the first type of approach can be found in \citeauthor{cheng:emnlp2015} \citeyear{cheng:emnlp2015}, where a recurrent neural network in which word embeddings were split into multiple sense vectors was proposed. The network was applied to paraphrase detection with positive results. 

In general, the evaluation of sense distinction models in the context of compositionality has often been evaluated on generic benchmarks, such as paraphrase detection. 
Despite the potential benefit in tasks such as question answering and information retrieval, there have been no attempts at integrating sense representations as components of neural compositional models.










\section{Conclusions}
\label{conclusion}

In this survey we have presented an extensive overview of semantically-grounded models for constructing distributed representations of meaning. Word embeddings have been shown to provide interesting semantic properties that can be applied to most language applications. However, these models tend to conflate different meanings into a single representation. Therefore, an accurate distinction of senses is often required for a deep understanding of lexical meaning. To this end, in this article we discuss models that learn representation for senses which are either directly induced from text corpora (i.e., unsupervised) or defined by external sense inventories (i.e., knowledge-based).

Some of these models have already proved effective in practise, but there is still much room for improvement. For example, even though semantically-grounded information is captured (to different degrees) by almost all models, common-sense reasoning has not yet been deeply explored. 
Also, most of these models have been tested on English only, whereas only a few have proposed models for other languages or attempted multilinguality. 
Finally, the integration of these theoretical models into downstream applications is the next step forward, as it is not clear now what the best integration strategy would be, and if a pre-disambiguation step is necessary. 
For instance, approaches such as the contextualized embeddings of \citeauthor{deepcontextual:2018} \citeyear{deepcontextual:2018} have shown a new possible direction in which senses are learned dynamically for each context, without the need for an explicit pre-disambiguation step. 

Although not exactly distributed representations of meaning, modelling relations in a flexible way is also another possible avenue for future work. Relations are generally modeled in works targeting knowledge-based completion. 
Moreover, a recent line of research has focused on improving relation embeddings with the help of text corpora \cite{toutanova2015representing,jameel2018relation,anke2018seven}, which paves the way for new approaches integrating these relations into downstream text applications.

From this perspective, the definition of sense and the correct paradigm 
is certainly still an open question. Do senses need to be discrete? Should they need to be tied to a knowledge resource or sense inventory? Should they be learned dynamically depending on the context? These are the questions that are yet to be explored according to the many studies on this topic. 
As also explained in our analysis, some approaches are more suited to certain applications or domains, without any clear general conclusion. These open questions are certainly still relevant and encourage further research on distributed representations of meaning, with many areas yet to be explored. \newline


\acks{The authors wish to thank the anonymous reviewers for their comments which helped improve the overall quality of this survey. The research of Jose Camacho-Collados is supported by ERC Starting Grant 637277.}



\vskip 0.2in
\bibliography{sample}
\bibliographystyle{theapa}

\end{document}